\documentclass[twocolumn]{svjour3}          
\smartqed  
\usepackage{color}
\usepackage{soul}
\usepackage{graphicx}
\usepackage{graphicx}
\usepackage{epsfig}
\usepackage{color}
\usepackage{pst-grad} 
\usepackage{pst-plot} 

\usepackage{amssymb}
\usepackage{amsmath}
\usepackage{cases}
\usepackage[ruled,vlined]{algorithm2e}
\usepackage{lineno}
\usepackage{xcolor}
\usepackage{marvosym}
\usepackage{stfloats}

\newcommand*{\LargerCdot}{\raisebox{-0.8ex}{\scalebox{1.8}{$\cdot$}}}

 \journalname{3D Research}
\begin{document}
\title{Construction of extended 3D field of views of the internal bladder wall surface: a proof of concept}
%
%
\author{Achraf Ben-Hamadou \and
        Christian Daul \and
        Charles Soussen
}
%
%
\institute{A. Ben-Hamadou \and C. Daul (\Letter) \and C. Soussen
\at
Universit\'e de Lorraine and CNRS. CRAN,
2 avenue de la For\^et de Haye,
F-54518 Vand{\oe}uvre-l\`es-Nancy, France\\
Tel.: +33 3 83 59 57 15\\
Fax: +33 3 83 59 56 44\\
\email{christian.daul@univ-lorraine.fr}  \\
\emph{Current affiliation:} A. Ben-Hamadou \\
             Valeo, Driving Assistance Research, 34 rue St-Andr\'e Z.I. des Vignes, 93012 Bobigny, France
}
\date{Received: 29 February 2016 / Revised: 25 May 2016 / Accepted: 30 May 2016}
%
%
\maketitle
\begin{abstract}
  3D extended field of views (FOVs) of the internal bladder wall
  facilitate lesion diagnosis, patient follow-up and treatment
  traceability. In this paper, we propose a 3D image mosaicing
  algorithm guided by 2D cystoscopic video-image registration for
  obtaining textured FOV mosaics. In this feasibility study, the
  registration makes use of data from a 3D cystoscope prototype
  providing, in addition to each small FOV image, some 3D points
  located on the surface. This proof of concept shows that textured
  surfaces can be constructed with minimally modified cystoscopes.
  The potential of the method is demonstrated on numerical and real
  phantoms reproducing various surface shapes. Pig and human bladder
  textures are superimposed on phantoms with known shape and
  dimensions. These data allow for quantitative assessment of the 3D
  mosaicing algorithm based on the registration of images simulating
  bladder textures.
\end{abstract}
\section{Introduction\label{sec:intro}}
In the United States, bladder cancer is the fourth and ninth most
widespread cancer among males and females, respectively, whereas in
Europe, approximately 50,000 people die from this disease each year
\cite{Alcaraz:2007}. Due to the high recurrence rate of $50\%$,
lifetime monitoring of patients is required after surgical removal of
cancer tumors \cite{Holzbeierlein:2004}. During a cystoscopy, a
flexible or rigid endoscope is inserted into the bladder through the
urethra, and a video of the epithelium is visualized on a screen. The
clinician (urologist or surgeon) scans the epithelium by maintaining
the distal tip of the cystoscope close to the internal organ wall. The
resulting highly resolved videos yield accurate visualization of the
wall through small field of view (FOV) images. However, neither
complete regions of interest (multifocal lesions, scares, etc.) nor
anatomical landmarks (urethra, ureters, air bubbles) are visible in
these images. Therefore, lesion diagnosis and follow-up may be
difficult and time-consuming. Extended FOV mosaics can be obtained by
superimposing the regions which can be jointly observed in small FOV
images. Not only do they facilitate lesion diagnosis and follow-up,
but they are also useful for data archiving (videos are highly
redundant data) and medical traceability (videos are not easy to
interpret by a clinician having not performed the acquisition).

\subsection{Bladder image mosaicing}
Bi-dimensional (2D) bladder image mosaicing algori\-thms were proposed
for the standard white-light modality \cite{Bergen13,Miranda2008} and
for the fluorescence modality \cite{Behrens:2008}. A complete
mosaicing algorithm consists of \emph{(i)} the registration of
consecutive images of the video sequence~\cite{Hernandez:2010},
\emph{(ii)} the correction of image misalignment in the mosaic leading
to bladder texture discontinuities \cite{Weibel:2012a}, and
\emph{(iii)} the contrast, intensity and/or color correction to handle
image blurring, shading effects and/or instrument viewpoint changes
\cite{Behrens:2010,Weibel2012b}. In \cite{Weibel:2012a}, it was shown
that mosaics covering large bladder surfaces (\emph{e.g.}, half of the
organ) can be computed. Although such 2D mosaics (see
Fig.~\ref{fig:1}) facilitate lesion diagnosis and follow-up, these
representations can be improved. Indeed, only the first image of the
mosaic has the original image resolution, the resolution of the other
images depending on the instrument trajectory and orientation.
Moreover, these images are strongly distorted during their placement
into the 2D map due to perspective changes of the endoscope. Both
image distortion and changes of resolution lead to mosaics with
spatially-varying visual quality. At last, urologists or surgeons
represent themselves the bladder in the three-dimensional (3D)
space. Mosaics in 3D (\emph{i.e.}, bladder wall surfaces with
superimposed 2D image textures) not only match this mental 3D
representation, but allow for a bladder texture representation without
any image distortion and with preserved resolution. Moreover, virtual
navigation inside the reconstructed organ part is possible after
clinical examinations using 3D mosaics. The interest of 2D and 3D
endoscopic image mosaicing in various medical fields is discussed in
the overview paper~\cite{Bergen16}.
\subsection{Previous work in 3D endoscopic data mosaicing}
In various fields of medical endoscopy, 3D mosaics are built following
three kinds of approaches.
%
%
%
%
%
%
\begin{figure}[t]
\centering
\psfig{figure=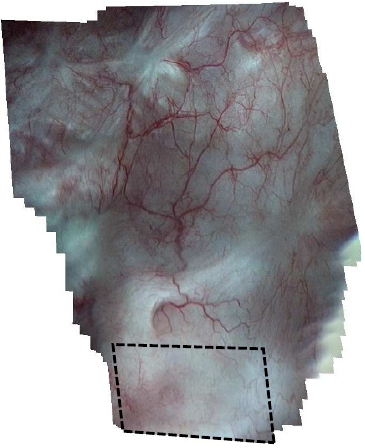, width=\linewidth}
\caption{Bladder mosaic (clinical data) obtained with the registration
  algorithm~\cite{Weibel:2012a} and using the
  algorithm~\cite{Weibel2012b} for correction of illumination
  discontinuities. The first image of the sequence is at the top
  right. From this image position, the endoscope first moves downwards
  in the bright regions (dashed bounding box) with weak
  textures. Then, its trajectory goes up towards the last image at the
  top left. Strong perspective and scale changes arise between the
  first and last images.}
\label{fig:1}
\end{figure}

In the first approach, analytically known surface shapes (representing
organs) are used both to limit the 2D registration errors and
visualize textured surfaces. In \cite{Behrens:2009a}, bladder
hemispheres were approximated by hemi\-cubes whose five sides
represent quasi planar bladder regions. A video sequence was then
partitioned into image groups, each group corresponding to some
hemicube side. The five reconstructed 2D sub-mosaics were then
projected onto the hemisphere to obtain a 3D bladder representation.
As for the inner esophagus wall~\cite{Carroll:2009}, the organ was
approximated by tubular shapes. The registration method aligns
endoscopic images using jointly the image colors and a mosaicing
surface of known geometry to reduce distortions. This algorithm
provides both the endoscope trajectory and a 2D mosaic that can be
projected onto the inner cylinder surfaces. Such approach based on 3D
prior models is well-suited to the esophagus which can indeed be
modelled by tubes, but is less appropriate for 3D bladder
mosaicing. Indeed, the bladder shape is hardly predictable, notably
since it is deformed by neighboring organs putting pressure on
it. Moreover, grouping images with respect to hemicube sides is
time-consuming and difficult.

In the second approach, no 3D prior information is imposed,
\emph{i.e.}, only video-images are taken into account. Different
passive vision methods were proposed based on shape from shading (SfS
\cite{Wu:2010}, bone surface reconstruction), structure from motion
(SfM \cite{Soper2012}, cystoscopy), shape from motion associated to
SfS (\cite{Kaufman:2008}, colon surface reconstruction),
SfM-factorization based on surgical instrument tracking
(\cite{Wu:2007}, surgery) or simultaneous localization and mapping
methods (SLAM \cite{Mountney2009}, laparoscopy for minimal invasive
surgery).
The lack of intensity variation in standard bladder images impedes
SfS. However, SfM or SLAM methods may be appropriate in urology. In
their 3D bladder mosaicing feasibility study~\cite{Soper2012}, Soper
\emph{et al} replaced the endoscope standardly used in urology by an
ultrathin fiber. The fiber was translated and rotated in such a way
that a complete bladder scan is performed. The acquired image sequence
was first used to find point correspondences in image pairs using
homologous feature points. Then, the 3D surface was constructed by
displacing a set of points initially located on a sphere.
Specifically, a bundle adjustment method was used to iteratively
determine both fiber tip trajectory and pig bladder phantom surface.
The main limitation of this approach for standard cystoscopy is that
salient texture points must be extracted from the video-images. In
large human bladder regions, textures may be missing or very weakly
contrasted for both healthy (see the bottom of the mosaic in
Fig.~\ref{fig:1}) and damaged tissues, \emph{e.g.}, due to scares. In
such images, SfM or SLAM techniques based on SIFT or SURF approaches
for extracting and matching textures thus suffer from a lack of
robustness (see \cite{Ali:2013,Hernandez:2010} for a detailed
discussion on feature based techniques in cystoscopy).
Feature point extraction and matching techniques is also difficult
because of illumination changes occurring between images, which are
due to strong perspective changes and/or the vignetting effect of
endoscopes. Another major limitation of feature-based methods is blur,
which is commonly encountered in standard cystoscopy images.

A less investigated approach in 3D endoscopy is based on active
vision, where surface points are reconstructed in the small FOV of
modified endoscopes. In \cite{Chan2003}, a two-channel endoscope was
used to reconstruct the inner surface of the mouth. The first channel
projects a structured-light pattern whereas the second acquires
video-images. Classical triangulation methods were used to reconstruct
3D points. Tests on phantoms have shown that submillimetre point
reconstruction accuracy can be reached with a small baseline of 2
mm. Another acquisition setup \cite{Penne2009} consists of a classical
CCD-camera and a time of flight (ToF) camera providing a depth map of
the FOV. A calibration procedure provides for each pixel both a color
and a 3D position with millimetre accuracy.
In \cite{Penne2009}, a two-channel endoscope (laparoscope) was used to
reconstruct points in a pig stomach. It is noticeable that such camera
pair can be used as well for single channel endoscopes such as
cystoscopes. Indeed, a ToF camera uses infrared signals that
do not interfere with the visible light of a color camera.
Shevchenko \emph{et al} proposed a structured-light approach for
obtaining extended 3D bladder surfaces~\cite{Shevchenko:2012}.
The prototype in~\cite{Shevchenko:2012} consists of an external
navigation system (which localizes the endoscope position in the
examination room) coupled with an endoscopic system projecting a grid
of points on the inspected surface. The 3D surface was built by
computing a 2D mosaic from the CCD camera images, and exploiting the
information provided by the external system. Knowing the endoscope
position and the 2D image correspondence, the 2D mosaic has to be
projected onto the reconstructed surface.
It is noticeable that in \cite{Agenant:2013}, the authors also
proposed a navigation system based method to document the position of
a lesion into the bladder during a first cystoscopy. The aim was not
3D image mosaicing, but rather to facilitate follow-up by performing
fast localization of lesions in upcoming cystoscopies.
In the conference paper \cite{Ben-Hamadou:2010}, we proposed the first
active vision based approach for 3D bladder mosaicing. Based on a
similar structured light principle as that described later in
\cite{Shevchenko:2012}, the proposed solution did however not require
any external navigation system for endoscope localization. The 3D
surface construction algorithm was based on 2D image registration and
on the reconstruction of 3D points in the coordinate system of the
endoscope camera. However, the proof of concept of 3D cystoscopy was
only partly established in \cite{Ben-Hamadou:2010}. The algorithm was
not tested using data (2D images and 3D points) acquired with the
structured light principle embedded in a cystoscope. The related data
were acquired for a large baseline (distance between the focal lengths
of the camera and the light projector) of about 150 mm. This approach
was not yet evaluated in the less favorable case of small baselines of
2 to 3 mm, like in cystoscopy. Furthermore, the image processing
algorithm in \cite{Ben-Hamadou:2010} needed to be improved. It was
based on a registration step, where the squared difference between
image grey-levels was minimized. This approach is not robust when
textures are weakly contrasted or with high inter- and intra-patient
variability. Such similarity measure is also not appropriate to the
illumination changes occurring in cystoscopic data.
%
%
%
%
\begin{figure*}[t]
\begin{tabular}{cc}
\hspace{-1mm}\includegraphics[width=0.45\textwidth]{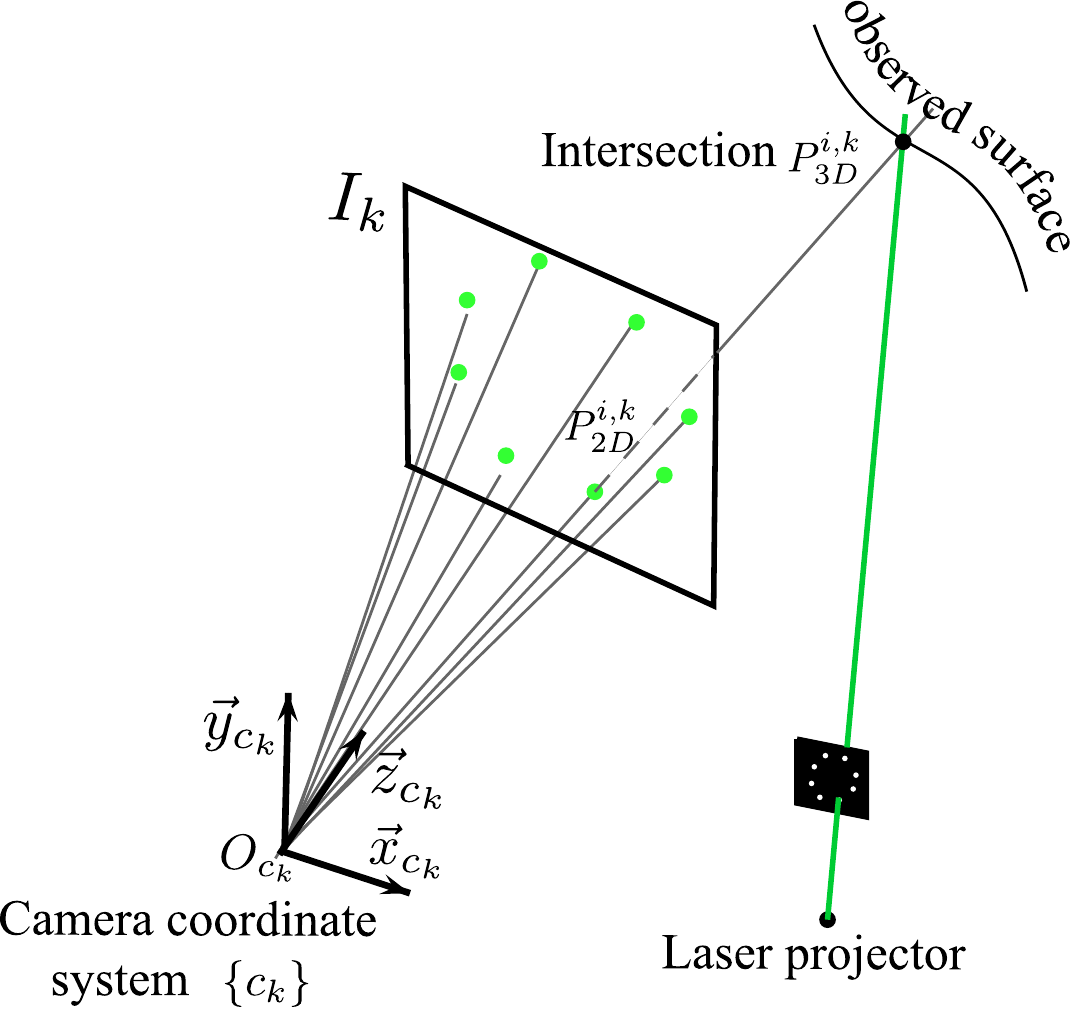}
&
\hspace{9mm}\includegraphics[width=0.45\textwidth]{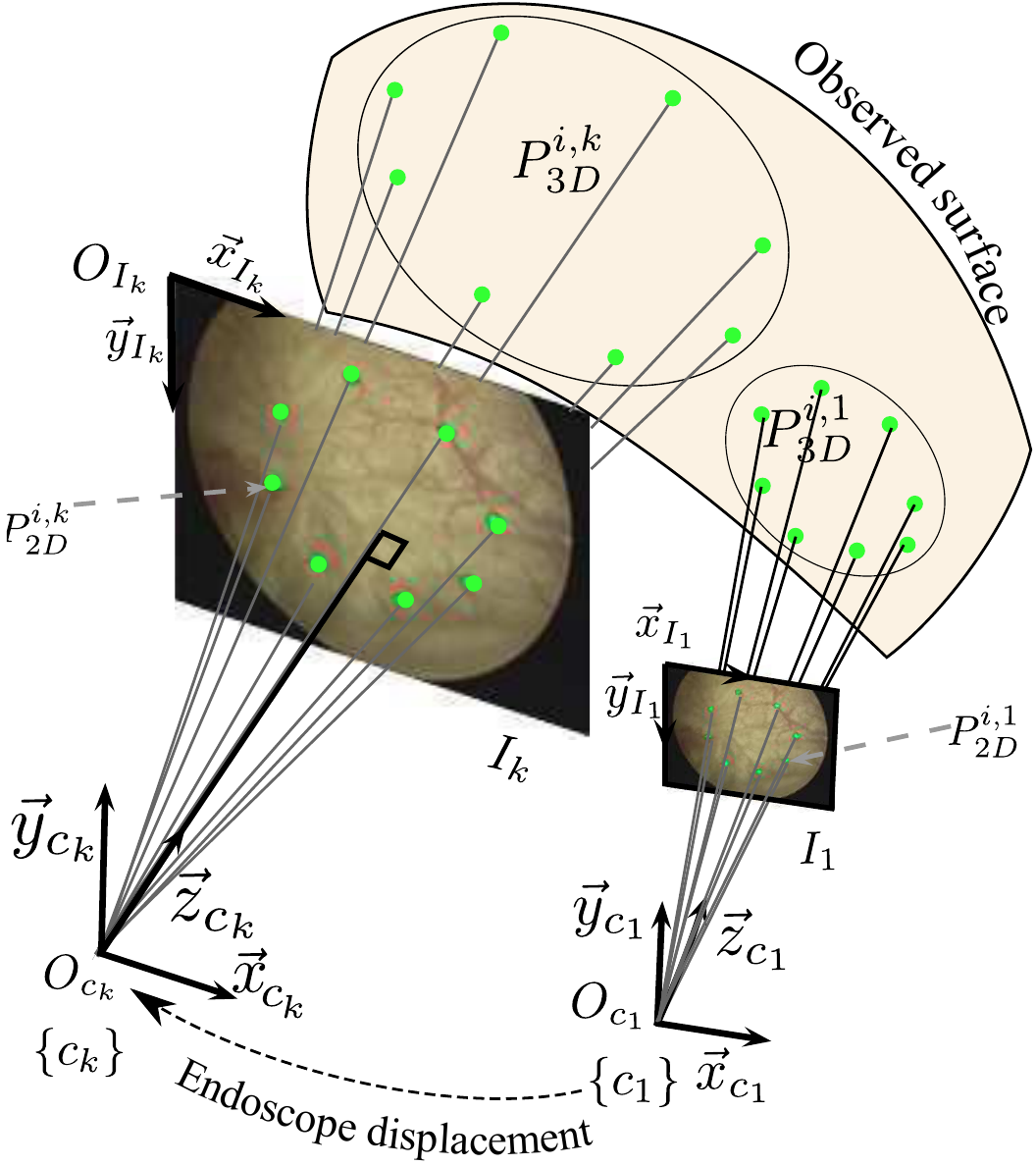}
\\
(a)& (b)
\end{tabular}
\caption{Acquisition principle~(a) and available data~(b) for 3D
  mosaicing. The acquisition principle is implemented using the
  prototype of Section~\ref{sec:ProtoResultsDiscussion}.\, (a) The
  $i$-th ``projector ray'' is shown in green. The corresponding
  ``camera ray'' is the black line originating from point
  $P_{3D}^{i,k}$ of the surface (intersection of the projector and
  camera rays) and passing through the optical center $O_{c_k}$ of the
  camera. $P_{2D}^{i,k}$ is the intersection of the camera ray with
  the plane supporting the acquired image $I_k$.\, (b) The available
  data are represented for the first and $k$-th acquisitions. The
  camera coordinate system $\{c_k$\} reads $(O_{c_k},
  \protect\overrightarrow{x_{c_k}}, \protect\overrightarrow{y_{c_k}},
  \protect\overrightarrow{z_{c_k}})$ and moves together with the
  prototype. The black line in bold represents the camera focal
  distance $f$, calibrated as described in \cite{Ben-Hamadou:2013}.}
\label{fig:2}
\end{figure*}

\subsection{Objectives\label{sec:objectives}}
The medical objective of the scanning fiber endoscope described in
\cite{Soper2012} was to design a procedure where bladder walls can be
fully acquired by nurses or ancillary care providers. This allows for
sparing the time of urologists who can post-operatively review the
bladder data. Such procedure is currently not standard but may be
complementary to the usual examination with flexible or rigid
cystoscopes. In the present paper, we aim to establish the proof of
concept of bladder mosaicing using 3D cystoscopes based on active
vision, which is a promising approach as shown
in~\cite{Shevchenko:2012}. In \cite{Soper2012}, the distal tip of the
fiber follows a spiral shaped trajectory and the distance from the
optical center to the bladder wall is in average larger than the
acquisition distance with cystoscopes. This allows for the use of the
SfM approach. This is no longer true with cystoscopy, since depth
disparity is often not guaranteed. Moreover, the lack of textures (as
in the bounding box of Fig. \ref{fig:1}) also complicates SfM
approaches, which require robust and accurate homologous point
correspondence between images.
In the present 3D bladder mosaicing feasibility study, an active
vision solution is chosen. As compared to \cite{Shevchenko:2012}, our
goal is to show that a 3D mosaicing algorithm guided by image
registration is useful for retrieving the endoscope displacement
between two acquisitions without using external navigation systems.
From the instrumentation viewpoint, we show that 3D bladder mosaicing
is feasible by simply modifying standard cystoscopes. Although the
proposed prototype cannot currently be used in clinical situation, our
3D point reconstruction approach can be implemented on cystoscopes.
Furthermore, it was shown that 3D point reconstruction is accurate
enough, even for small baselines of point triangulation
\cite{Ben-Hamadou:2013,Chan2003}.

From the image processing viewpoint, we propose a 3D image mosaicing
algorithm for which no assumption has to be made regarding the
endoscope trajectory. The unknown endoscope displacement between
consecutive image acquisitions is a general combination of a 3D
rotation and a 3D translation. Although real-time mosaicing can be of
interest, the computation time is not the most critical point in
urology. As argued in \cite{Miranda2008,Weibel:2012a}, a mosaicing
time of some tens of minutes is acceptable for a second (finer)
diagnosis carried out to confirm or modify the first diagnosis during
the real-time visualization of the small FOV video-images. Real-time
is neither a critical factor for lesion follow-up because the lesion
evolution assessment is performed with data from a cystoscopy
scheduled some weeks or months after the cystoscopy in progress.

The paper is organized as follows. Section \ref{sec:AcquisitionData}
introduces the data acquisition principle (for readability reasons,
the description of the prototype implementing this measurement
principle is deferred to Section~\ref{sec:ProtoResultsDiscussion}).
Section~\ref{sec:3Dmosaicing} details the 3D mosaicing algorithm.
In Section~\ref{sec:ProtoResultsDiscussion}, the algorithm is
tested on phantoms with various shapes allowing for a validation
of the algorithm with data acquired by the proposed 3D cystoscope prototype.
In Section~\ref{sec:SimulatedDataResultsDiscussion}, simulated data allow
for qualitative evaluation on more realistic bladder data, in terms of
both 3D surface shape and bladder texture.
Section \ref{sec:3DRegistrationError} exhibits the 3D registration
errors obtained for both synthetic data and data related to the
prototype. Concluding remarks and perspectives are given in
Section~\ref{sec:Conclusion}.
\section{Acquisition principle and available data \label{sec:AcquisitionData}}
The principle of our laser based active vision system is sketched in
Fig.~\ref{fig:2}(a). The prototype is equipped with a color camera and
a structured laser light projector. The basic idea is to project laser
rays, such as the green line of Fig.~\ref{fig:2}(a), onto the bladder
epithelial surface. Then, each projector ray is being reflected by the
surface. This generates a so-called ``camera ray'' (the black line of
Fig.~\ref{fig:2}(a)) passing through the green laser dots visible in
image $I_k$. In Fig.~\ref{fig:2}(a), the quadrangle with black
background corresponds to diffractive optics, \emph{i.e.}, holographic
binary phase lens, designed to project eight laser rays. The latter
are located on a cone originating from the projector optical
center. Each projector ray passes through one of the eight white dots
of the diffractive optics. For readability matters, only one ray is
shown in Fig.~\ref{fig:2}(a).

The calibration procedure in \cite{Ben-Hamadou:2013} is used to
compute the intrinsic and extrinsic camera and projector
parameters. The parameters of the projector ray equations are computed
in the camera coordinate system $(O_{c_k}, \overrightarrow{x_{c_k}},
\overrightarrow{y_{c_k}}, \overrightarrow{z_{c_k}})$, referred to as
$\{c_k$\}, with $O_{c_k}$ the optical center of the camera. Because
the projector and camera are rigidly fixed together, the equations of
the projector rays in $\{c_k$\} do not depend on $k$. The equations of
the camera rays (passing all through $O_{c_k}$) are also computed in
$\{c_k$\} given the location $P_{2D}^{i,k}$ of the green dots in image
$I_k$ ($i$ stands for the index of the green dots). The green dots can
be easy segmented since the hue of the bladder is always reddish or
orange, see the real images of Figs.~\ref{fig:1} and
\ref{fig:2}(b). The $i$-th laser point on the bladder surface
($P_{3D}^{i,k}$) is the intersection of the projector ray with its
corresponding camera ray.

The data available for 3D mosaicing are related to a video-sequence
with $K$ frames. Each acquisition $k \in \{1,\ldots,K\}$ corresponds
to a cystoscope viewpoint and yields: \emph{(i)}~an image $I_k$;
\emph{(ii)}~eight laser dots $P_{2D}^{i,k}=(x_{2D}^{i,k},
y_{2D}^{i,k})^T$ given in the coordinate system $(O_{I_k},
\overrightarrow{x_{I_k}},\\ \overrightarrow{y_{I_k}}$) of the plane
supporting $I_k$; \emph{(iii)}~eight laser points
$P_{3D}^{i,k}=(x_{3D}^{i,k}, y_{3D}^{i,k}, z_{3D}^{i,k})^T$ given in
the camera coordinate system $\{c_k$\} (see Fig.~\ref{fig:2}(b)). The
barrel distortions in images $I_k$ are corrected using the known
intrinsic camera parameters
\cite{Ben-Hamadou:2013}. The laser rays ($i \in \{1,\ldots,M\}$ with
$M=8$) are projected in the periphery of the circular FOV of the
images to minimize the loss of bladder texture and to allow for
diagnosis. Although only eight $P_{3D}^{i,k}$ points are available per
viewpoint, the video acquisition leads to a large number of points
overall due to the high acquisition speed (25 images/s) and the slow
endoscope displacements (some mm/s).

The algorithm proposed in Section~\ref{sec:3Dmosaicing} aims at
placing all points $P_{3D}^{i,k}$ in a common coordinate system, and
then computing a polygonal mesh representing the smooth bladder surface,
without corners nor sharp edges. The textures of images $I_k$ will
finally be projected onto the polygonal mesh.

\section{Construction of three-dimensional textured surfaces \label{sec:3Dmosaicing}}
The proposed 3D mosaicing method takes as inputs a set of
distortion-corrected images $I_k$ and a set of $M=8$ points per
viewpoint $k$. Each $P_{3D}^{i,k}$ point is given in the local
(moving) coordinate system $\{c_k$\} of the camera. Moreover, their 2D
projections $P_{2D}^{i,k}$ in the $k$-th image of the video-sequence
($i=1,\ldots,M$) are known as well. The algorithm for reconstructing
the $P_{3D}^{i,k}$ points from the knowledge of their projections
$P_{2D}^{i,k}$ can be found in~\cite{Ben-Hamadou:2013}.

The mosaicing algorithm requires some geometric prerequisites, which
are presented now.

\subsection{Geometrical considerations \label{sec:geometry}}
The first geometrical transformation involves the intrinsic camera
parameters linking the 3D camera coordinate system $\{c_k$\} to the 2D
image plane coordinate system $(O_{I_k}, \overrightarrow{x_{I_k}},
\overrightarrow{y_{I_k}}$). In~\eqref{eq:matricePerspective}, the
perspective projection matrix ${\mathbf{K}}$ is fully defined from the
knowledge of the projection $(u, v)$ of the camera optical center in
image $I_k$, the focal length $f$, and the CCD-sensor size $l_x$ and
$l_y$ along the $\overrightarrow{x_{I_k}}$ and
$\overrightarrow{y_{I_k}}$ axes (see Fig.~\ref{fig:2}(b)):
\begin{align}
\left[
\setlength\arraycolsep{3.0pt}
\begin{array}{c}  P_{2D}^{i,k}  \\ 1
\end{array}\right]
=\,
\frac{1}{z_{3D}^{\,i,k}} \, {\mathbf{K}} \, P_{3D}^{i,k} \,\,\,
\textrm{with} \,\,\,\,
\label{eq:matricePerspective}
{\mathbf{K}}=
\left[
\setlength\arraycolsep{1.0pt}
\begin{array}{ccc}
f/l_x & 0             & u     \\
   0         & f/l_y & v  \\
  0         &  0            & 1
\end{array}\right].
\end{align}

The second transformation is a 3D rigid transformation linking
consecutive camera viewpoints. It is parameterized by a $3\times 4$
matrix $\mathbf{T}_{3D}^{k-1,k}$. $\mathbf{T}_{3D}^{k-1,k}$ is
composed of a $3\times 3$ rotation matrix $\mathbf{R}^{k-1,k}$ and a
$3\times 1$ translation vector $D^{k-1,k}$:
\begin{align}
\hat{P}_{3D}^{i,k-1}
=
\mathbf{T}_{3D}^{k-1,k} \,
\!\!\left[\begin{array}{c} P_{3D}^{\,i,k}\\1
\end{array}\right],
\label{eq:matriceTrigidea}%
\end{align}
with
\begin{align}
\mathbf{T}_{3D}^{k-1,k}
=
\left[
\setlength\arraycolsep{2pt}
\begin{array}{c|c}
\mathbf{R}^{k-1,k} & D^{k-1,k}
\end{array}\right].
\label{eq:matriceTrigideb}%
\end{align}
The coefficients of the rotation matrix are defined from the
well-known Euler angles, denoted by $\theta_1^{k-1,k}$,
$\theta_2^{k-1,k}$ and $\theta_3^{k-1,k}$. The so-called
``$x$-convention'' is used for these angles: $\theta_1^{k-1,k} \in
[0,2\pi]$ is the angle around the $\overrightarrow{z_{c_k}}$-axis,
$\theta_2^{k-1,k}\in [0, \pi]$ is around the new (rotated)
$\overrightarrow{x_{c_k}}^{\,'}$-axis, while $\theta_3^{k-1,k}\in
[0,2\pi]$ is around the rotated
$\overrightarrow{z_{c_k}}^{\,''}$-axis. Superscripts $'$ and $''$
respectively refer to the new location of axes
$\overrightarrow{x_{c_k}}$ and $\overrightarrow{z_{c_k}}$ after the
first and second rotations. The matrix $\mathbf{T}_{3D}^{k-1,k}$
displaces point $P_{3D}^{i,k}$ in $\{c_k\}$ to
$\hat{P}_{3D}^{i,k-1}=(\hat{x}_{3D}^{i,k-1}, \hat{y}_{3D}^{i,k-1},
\hat{z}_{3D}^{i,k-1})^T$ in $\{c_{k-1}$\}. It is worth noticing that
$\hat{P}_{3D}^{i,k-1}$ are points brought wi\-thin $\{c_{k-1}$\}
whereas $P_{3D}^{i,k-1}$ are ``data points'' reconstructed in
$\{c_{k-1}\}$ using the calibration parameters and the segmented dots
$P_{2D}^{i,k-1}$. Thus, $P_{3D}^{i,k-1}$ and $\hat{P}_{3D}^{i,k-1}$
represent different laser points.

The third relationship required by the proposed me\-thod is a
homography linking two consecutive images $I_{k}$ and $I_{k-1}$. This
2D relationship is established under several working
assumptions. First, the distal tip of the cystoscope is usually close
to the inner bladder epithelium. The FOV being very limited, the
surfaces viewed in the images are usually small and are assumed to be
planar. Second, the bladder is filled by an isotonic saline solution
which rigidifies the bladder wall. The acquisition speed being high
(25 image/s) and the endoscope displacement speed low (some mm/s), we
assume that the common surface part viewed in images $I_{k-1}$ and
$I_{k}$ does not warp itself between two consecutive
acquisitions. Based on these working assumptions, the homography is
well-suited to model the dependence between $I_{k}$ and
$I_{k-1}$. This choice was practically validated for numerous
algorithms for 2D/2D registration of bladder images: feature-based
methods in \cite{Behrens:2009a} (fluorescence modality) and
\cite{Bergen13,Soper2012} (white-light modality), the graph-cut method
in \cite{Weibel:2012a} and two optical flow methods in
\cite{Ali16b,Ali16a}. The homography matrix is defined by:
\begin{align}
\mathbf{T}_{2D}^{k-1,k} = \left[\begin{array}{ccc}
 \underbrace{\alpha^{k-1,k}\cos\varphi^{k-1,k}}_{a^{k-1,k}_{1,1}} &
 \underbrace{-s_x^{k-1,k}\sin\varphi^{k-1,k}}_{a^{k-1,k}_{1,2}} &
 \underbrace{t^{k-1,k}_{x, 2D}}_{a^{k-1,k}_{1,3}}  \\
 \underbrace{s_y^{k-1,k}\sin\varphi^{k-1,k}}_{a^{k-1,k}_{2,1}} &
 \underbrace{\alpha^{k-1,k}\cos\varphi^{k-1,k}}_{a^{k-1,k}_{2,2}} &
 \underbrace{t^{k-1,k}_{y, 2D}}_{a^{k-1,k}_{2,3}}  \\
 a_{3,1}^{k-1,k} &
 a_{3,2}^{k-1,k} & 1
\end{array}\right]
\label{eq:matriceHomography}
\end{align}
where the parameters $\alpha$, $\varphi$, $(s_x, s_y)$ and $(t_{x,2D},
t_{y,2D})$ respectively denote the scale factor, in-plane rotation,
shearing and 2D translation changes. In the following and for
simplicity reasons, the matrix $\mathbf{T}_{2D}^{k-1,k}$ will be
simply described by means of its coefficients $a^{k-1,k}_{r,s}$.

Image $I_k$ is registered with $I_{k-1}$ when
$\mathbf{T}_{2D}^{k-1,k}$ superimposes homologous points $P_{2D}^{k}$
and $P_{2D}^{k-1}$ of both images, \emph{i.e.}, when
\begin{align}
\left[\begin{array}{c}  P_{2D}^{k-1} \\ 1
\end{array}\right]
&\!\!=\!
\frac{1}{\beta^{k-1,k}}  \,\, \mathbf{T}_{2D}^{k-1,k}
\left[\begin{array}{c}  P_{2D}^{k} \\ 1
\end{array}\right].
\label{eq:PointsHomologues}
\end{align}
%
%
The normalizing factor $\beta^{k-1, k}$ deduces from the perspective
parameters $a_{3,1}^{k-1,k}$ and $a_{3,2}^{k-1,k}$:
\begin{align}
\beta^{k-1,k}=[a_{3,1}^{k-1,k},a_{3,2}^{k-1,k}]^TP_{2D}^{k}+1.
\label{eq:beta}
\end{align}
Let us stress that \eqref{eq:PointsHomologues} is met for all
homologous texture pixels of images $I_k$ and $I_{k-1}$
($P_{2D}^{k-1}$ and $P_{2D}^{k}$), but not for the laser points
$P_{2D}^{i,k}$ and $P_{2D}^{i,k-1}$. Indeed, $P_{2D}^{i,k}$ and
$P_{2D}^{i,k-1}$ are not homologous points because the active vision
system is moving from viewpoint $k-1$ to viewpoint $k$. Nevertheless,
\eqref{eq:PointsHomologues} will be exploited to estimate the
parameters of a candidate homography $\mathbf{T}_{2D}^{k-1,k}$
corresponding to a candidate $\mathbf{T}_{3D}^{k-1,k}$
transformation. Moreover, the quality of image superimposition will be
assessed based on \eqref{eq:PointsHomologues}, applied to all
homologous texture pixels of $I_{k-1}$ and $I_k$.

\setcounter{equation}{12}
\begin{table*}[bp]
\begin{equation}
~~~~~~~~~~~~~~~~~~~~~~~~~~~~~\left[
\setlength\arraycolsep{8pt}
\begin{array}{cccccccc}
      x_{2D}^{1,k} & y_{2D}^{1,k} & 1 & 0 & 0 & 0 &
      -c^{k-1,k}_{1,1}x_{2D}^{1,k} & -c^{k-1,k}_{1,1}y_{2D}^{1,k} \\[.2cm]
      0 & 0 & 0 & x_{2D}^{1,k} & y_{2D}^{1,k} & 1 &
      -c^{k-1,k}_{2,1}x_{2D}^{1,k} & -c^{k-1,k}_{2,1}y_{2D}^{1,k} \\
      \vdots &  \vdots & \vdots & \vdots & \vdots & \vdots & \vdots &
      \vdots \\
      x_{2D}^{M,k} & y_{2D}^{M,k} & 1 & 0 & 0 & 0 &
      -c^{k-1,k}_{1,M}x_{2D}^{M,k} & -c^{k-1,k}_{1,M}y_{2D}^{M,k} \\[.2cm]
      0 & 0 & 0 & x_{2D}^{M,k} & y_{2D}^{M,k} & 1 &
      -c^{k-1,k}_{2,M}x_{2D}^{M,k} & -c^{k-1,k}_{2,M}y_{2D}^{M,k} \\
    \end{array}\right]
  \left[
\setlength\arraycolsep{0pt}
\begin{array}{c}
      a_{1,1}^{k-1,k} \\[.1cm] a_{1,2}^{k-1,k} \\[.1cm] a_{1,3}^{k-1,k} \\[.1cm] a_{2,1}^{k-1,k} \\[.1cm]
      a_{2,2}^{k-1,k} \\[.1cm] a_{2,3}^{k-1,k} \\[.1cm] a_{3,1}^{k-1,k} \\[.1cm] a_{3,2}^{k-1,k}
    \end{array}\right]
  =
  \left[
\setlength\arraycolsep{0pt}
\begin{array}{c}
      c_{1,1}^{k-1,k} \\[.2cm]
      c_{2,1}^{k-1,k} \\
      \vdots \\
      c_{1,M}^{k-1,k} \\[.2cm]
      c_{2,M}^{k-1,k}
    \end{array}\right]
  \label{eq:Lien2D3DMatricielFinal}
\end{equation}
\end{table*}
\setcounter{equation}{6}

\subsection{Consecutive viewpoint registration\label{sec:registration}}
\subsubsection{Principle of the registration algorithm}
Let us define the displaced projected laser points
$\hat{P}_{2D}^{i,k-1}$ as the 2D projections of the laser points
$P_{3D}^{i,k}$, displaced from $\{c_{k}$\} to $\{c_{k-1}$\}.
Combining~\eqref{eq:matricePerspective} and~\eqref{eq:matriceTrigidea}
yields:
\begin{align}
  \setlength\arraycolsep{0.0pt}
  \left[\begin{array}{c}
      \hat{P}_{2D}^{i,k-1} \\ 1
    \end{array}\right] & = \frac{1}{ \hat{z}_{3D}^{i,k-1}}\;
   {\mathbf{K}} \mathbf{T}_{3D}^{k-1,k} \left[
     \begin{array}{c} P_{3D}^{i,k}\\ 1
     \end{array}\right].
   \label{eq:PointsTransforme}
\end{align}
Now, consider an ideal (and searched) $\mathbf{T}_{3D}^{k-1,k}$
transformation, whose related homography $\mathbf{T}_{2D}^{k-1,k}$
superimposes the pixels of $I_k$ and their homologous pixels in
$I_{k-1}$. Such $\mathbf{T}_{2D}^{k-1,k}$ transformation not only
superimposes homologous texture pixels, but also relates each laser
point $P_{2D}^{i,k}$ to its displaced instance $\hat{P}_{2D}^{i,k-1}$:
\begin{align}
\setlength\arraycolsep{0.0pt}
  \left[\begin{array}{c} \hat{P}_{2D}^{i,k-1} \\ 1
     \end{array}\right]  = \frac{1}{\beta^{k-1,k}_i}\:\mathbf{T}_{2D}^{k-1,k}
   \left[\begin{array}{c} P_{2D}^{i,k} \\ 1
     \end{array}\right],
   \label{eq:PointsHomologuesModif}
\end{align}
%
where the factor $\beta^{k-1,k}$ appearing in
\eqref{eq:PointsHomologues} has been replaced by $\beta^{k-1,k}_i$ to
stress the dependence upon $i$.
Combining~\eqref{eq:PointsTransforme} and
\eqref{eq:PointsHomologuesModif} yields:
\begin{align}
\setlength\arraycolsep{0.0pt}
\frac{1}{\beta^{k-1,k}_i}\;\mathbf{T}_{2D}^{\,k-1,k}\left[
\begin{array}{c}
P_{2D}^{\,i,k}\\1
\end{array}\right]
&=\frac{1}{\hat{z}_{3D}^{\,i,k-1}}\,{\mathbf{K}}\;\mathbf{T}_{3D}^{k-1,k}
\left[\begin{array}{c} P_{3D}^{\,i,k}\\1
\end{array}\right]
\label{eq:Lien2D3DMatriciel}
\end{align}
which involves the 2D and 3D laser points related to viewpoint $k$ only.

For a given candidate $\mathbf{T}_{3D}^{k-1,k}$,
$\hat{z}_{3D}^{i,k-1}$ is known from the last equation
of~\eqref{eq:PointsTransforme} and the knowledge of $P_{3D}^{i,k}$.
Similarly, $\beta^{k-1,k}_i$ can be deduced from
$\mathbf{T}_{2D}^{k-1,k}$ using~\eqref{eq:beta}. Therefore,
$\mathbf{T}_{3D}^{k-1,k}$ and $\mathbf{T}_{2D}^{k-1,k}$ are the only
unknowns in~\eqref{eq:Lien2D3DMatriciel}. As we will see,
$\mathbf{T}_{2D}^{k-1,k}$ can actually be deduced from
$\mathbf{T}_{3D}^{k-1,k}$ and the knowledge of the laser points.

The proposed registration scheme is designed in such a way that some
similarity measure (to be detailed) between both images $I_k$ and
$I_{k-1}$ is maximized with respect to the $\mathbf{T}_{3D}^{k-1,k}$
parameters under the constraint formulated
in~\eqref{eq:Lien2D3DMatriciel}. Each iteration of the optimization
algorithm includes four steps related to the update and evaluation of
the candidate $\mathbf{T}_{3D}^{k-1,k}$:
\begin{enumerate}
\item Update the rigid transformation $\mathbf{T}_{3D}^{k-1,k}$.
\item Compute matrix $\mathbf{T}_{2D}^{k-1,k}$ induced by
  $\mathbf{T}_{3D}^{k-1,k}$ using \eqref{eq:Lien2D3DMatriciel} and the
  location of the $M$ laser points $(P_{3D}^{i,k},P_{2D}^{i,k})$.
\item Superimpose $I_{k-1}$ with $\mathbf{T}_{2D}^{k-1,k}(I_{k})$,
  which denotes the transformation of image $I_k$ using the computed
  homography $\mathbf{T}_{2D}^{k-1,k}$; see \eqref{eq:PointsHomologues}.
\item Evaluate the similarity measure between images $I_{k-1}$ and
  $\mathbf{T}_{2D}^{k-1,k}(I_{k})$ (the pixels of both images
  corresponding to laser dots are not taken into account).
\end{enumerate}
This process terminates when the similarity measure between the
registered images is maximal. This algorithm simultaneously provides
the homography $\mathbf{T}_{2D}^{k-1,k}$ and the rigid transformation
$\mathbf{T}_{3D}^{k-1,k}$ linking $\{c_{k}$\} to $\{c_{k-1}$\}.
Hereafter, we detail how $\mathbf{T}_{2D}^{k-1,k}$ is computed from
$\mathbf{T}_{3D}^{k-1,k}$. Then, we define the similarity measure used
for assessing the 2D image superimposition quality. Finally, an
overview of the optimization algorithm is given
(Algorithm~\ref{alg:transformationDetermination}).

\subsubsection{Matrix relationship between $\mathbf{T}_{2D}^{k-1,k}$ and $\mathbf{T}_{3D}^{k-1,k}$\label{sec:matrixrelationship}}
For a point correspondence $(P_{3D}^{i,k}, P_{2D}^{i,k})$,
\eqref{eq:Lien2D3DMatriciel} is a system of three
equations. Rearranging the first two equations and replacing
$\beta^{k-1,k}_i$ and $\hat{z}_{3D}^{\,i,k-1}$ by their expressions,
we obtain:
\begin{align}
\left\{
\begin{array}{ccc}
\frac{\mathbf{T}_{2D,1\LargerCdot}^{k-1,k}\left[
\setlength\arraycolsep{1.2pt}
\begin{array}{c}P_{2D}^{i,k}\\1\end{array}\right]}
     {\mathbf{T}_{2D,3\LargerCdot}^{k-1,k}\left[\begin{array}{c}P_{2D}^{i,k}\\1\end{array}\right]}
     & =
\frac{
{\mathbf{K}}_{1\LargerCdot}\;
         \mathbf{T}_{3D}^{k-1,k}\left[
\setlength\arraycolsep{1.2pt}
\begin{array}{c}P_{3D}^{i,k}\\1\end{array}\right]}
     {\mathbf{T}_{3D,3\LargerCdot}^{k-1,k}\left[
\setlength\arraycolsep{1.2pt}
\begin{array}{c}P_{3D}^{i,k}\\1\end{array}\right]}
\\[1.0cm]
\frac{\mathbf{T}_{2D,2\LargerCdot}^{k-1,k}\left[
\setlength\arraycolsep{1.2pt}
\begin{array}{c}P_{2D}^{i,k}\\1\end{array}\right]}
     {\mathbf{T}_{2D,3\LargerCdot}^{k-1,k}\left[
\setlength\arraycolsep{1.2pt}
\begin{array}{c}P_{2D}^{i,k}\\1\end{array}\right]}
     & =
\frac{
{\mathbf{K}}_{2\LargerCdot}\;
         \mathbf{T}_{3D}^{k-1,k}\left[
\setlength\arraycolsep{1.2pt}
\begin{array}{c}P_{3D}^{i,k}\\1\end{array}\right]}
     {\mathbf{T}_{3D,3\LargerCdot}^{k-1,k}\left[
\setlength\arraycolsep{1.2pt}
\begin{array}{c}P_{3D}^{i,k}\\1\end{array}\right]}
\end{array}
\right.
\label{eq:Lien2D3DNonMatriciel1}
\end{align}
where the subscripts $1\LargerCdot$, $2\LargerCdot$ and $3\LargerCdot$
respectively index the first, second and third matrix rows and all
column indices. Denoting by $c^{k-1,k}_{1,i}$ and $c^{k-1,k}_{2,i}$
the right-hand sides in~\eqref{eq:Lien2D3DNonMatriciel1},
\eqref{eq:Lien2D3DNonMatriciel1} rewrites:
\begin{align}
\left\{\begin{array}{ccc}
\mathbf{T}_{2D,1\LargerCdot}^{k-1,k}\left[\begin{array}{c}P_{2D}^{i,k}\\1\end{array}\right]
& = &
c^{k-1,k}_{1,i}\;\mathbf{T}_{2D,3\LargerCdot}^{k-1,k}\left[\begin{array}{c}P_{2D}^{i,k}\\1\end{array}\right]
\\[20pt]
\mathbf{T}_{2D,2\LargerCdot}^{k-1,k}\left[\begin{array}{c}P_{2D}^{i,k}\\1\end{array}\right]
& = &
c^{k-1,k}_{2,i}\;\mathbf{T}_{2D,3\LargerCdot}^{k-1,k}\left[\begin{array}{c}P_{2D}^{i,k}\\1\end{array}\right]
\end{array} \right. \label{eq:Lien2D3DNonMatriciel2}
\end{align}
with
\begin{align*}
c^{k-1,k}_{1,i}  & = \left (
\textbf{K}_{1\LargerCdot}\;
\mathbf{T}_{3D}^{k-1,k}\left[
\setlength\arraycolsep{0.0pt}
\begin{array}{c}P_{3D}^{i,k}\\1\end{array}\right]
\right )/
\left (\mathbf{T}_{3D,3\LargerCdot}^{k-1,k}\left[
\setlength\arraycolsep{0.0pt}
\begin{array}{c}P_{3D}^{i,k}\\1\end{array}\right]
\right ) \\
c^{k-1,k}_{2,i} & = \left (
\textbf{K}_{2\LargerCdot}\;
\mathbf{T}_{3D}^{k-1,k}\left[
\setlength\arraycolsep{0.0pt}
\begin{array}{c}P_{3D}^{i,k}\\1\end{array}\right]\right )
/
\left (
{\mathbf{T}_{3D,3\LargerCdot}^{k-1,k}\left[
\setlength\arraycolsep{0.0pt}
\begin{array}{c}P_{3D}^{i,k}\\1\end{array}\right]}\right )
\end{align*}
Replacing the points $P_{3D}^{i,k}$ and $P_{2D}^{i,k}$ by their
coordinates, \eqref{eq:Lien2D3DNonMatriciel2} yields a linear system
of two equations involving the coefficients $a_{r,s}^{k-1,k}$
in~\eqref{eq:matriceHomography}:
\begin{equation}
\left\{
\setlength\arraycolsep{1pt}
\begin{array}{l}
x_{2D}^{i,k}a_{1,1}^{k-1,k}+ y_{2D}^{i,k}a_{1,2}^{k-1,k} +
a_{1,3}^{k-1,k}- \ldots \medskip
\\
\hspace*{.2cm}c^{k-1,k}_{1,i}x_{2D}^{i,k}a_{3,1}^{k-1,k}  -
c^{k-1,k}_{1,i}y_{2D}^{i,k}a_{3,2}^{k-1,k}  = c^{k-1,k}_{1,i} \bigskip\\
x_{2D}^{i,k}a_{2,1}^{k-1,k} +y_{2D}^{i,k}a_{2,2}^{k-1,k} + a_{2,3}^{k-1,k}- \ldots\medskip \\
\hspace*{.2cm}c^{k-1,k}_{2,i}x_{2D}^{i,k}a_{3,1}^{k-1,k}-c^{k-1,k}_{2,i}y_{2D}^{i,k}a_{3,2}^{k-1,k}
=c^{k-1,k}_{2,i}

\end{array}
\right.
\label{eq:Lien2D3DNonMatriciel3}
\end{equation}
Both equations are valid for all laser points $P^{i,k}_{3D}$ of
viewpoint $k$. This yields a linear system of $2M$ equations with
eight unknown parameters $a_{r,s}^{k-1,k}$ related to transformation
$\mathbf{T}_{2D}^{k-1,k}$: see~\eqref{eq:Lien2D3DMatricielFinal}.
For our prototype, $M=8$, so \eqref{eq:Lien2D3DMatricielFinal} is an
over-determined system of 16 equations. Notice that when $M \geq 4$,
the homography $\mathbf{T}_{2D}^{k-1,k}$ induced by
$\mathbf{T}_{3D}^{k-1,k}$ is unique since there are $2M\geq 8$
equations in~\eqref{eq:Lien2D3DMatricielFinal}.
\setcounter{equation}{13}
%
%
\begin{algorithm*}
\caption{Data registration of viewpoints $k$ and $k-1$.}
\label{alg:transformationDetermination}
\KwIn{\begin{itemize}{}
       \item Source image $I_k$ and target image $I_{k-1}$.
       \item $M$ laser points of viewpoint $k$ given in 3D and 2D:
         $\{P_{3D}^{i,k}$, $P_{2D}^{i,k}\}|_{i\in \{1,\ldots,M\}}$.
       \item Initial simplex given in~\eqref{eq:initialSimplex}, and
       related similarity values $\varepsilon(V_e)$, $e \in \{1,\ldots,7\}$.
     \end{itemize}}
\KwOut{Geometrical transformation linking viewpoints $k$ and $k-1$:
     \begin{itemize}{}
     \item homography matrix $\mathbf{T}_{2D}^{k-1,k}$.
     \item translation vector $D^{k-1,k}$ and angles
       $\theta_1^{k-1,k}$, $\theta_2^{k-1,k}$ and $\theta_3^{k-1,k}$
       induced by the estimated $3\times 4$ rigid transformation
       matrix $\mathbf{T}_{3D}^{k-1,k}$.
     \end{itemize}}
    \Repeat{the variation of the simplex volume leads to a mean displacement
      value of $I_{k}$ on $I_{k-1} < 0.1$ pixel}
    {
      \underline{\em Step 1}: Replace the vertex of smallest
      $\varepsilon(V_e)$ value by a new vertex according to the
      symmetry rules of the simplex algorithm.
      Then, update the entries of $\mathbf{T}_{3D}^{k-1,k}$
        (see~\eqref{eq:matriceTrigideb}) from the knowledge of
      $\theta_1^{k-1,k}$, $\theta_2^{k-1,k}$, $\theta_3^{k-1,k}$,
      and of the 3D translation vector $D^{k-1,k}$ for the new
      vertex\;
      \underline{\em Step 2}: Given~\eqref{eq:Lien2D3DMatricielFinal}
      and the $M$ points $(P_{3D}^{i,k}, P_{2D}^{i,k})$, compute the 8
      parameters of the homography $\mathbf{T}_{2D}^{k-1,k}$ induced
      by the rigid transformation $\mathbf{T}_{3D}^{k-1,k}$ determined
      in {\em Step 1}\;
      \underline{\em Step 3}: Superimpose $I_k$ on $I_{k-1}$
      using~\eqref{eq:PointsHomologues} and the homography
      $\mathbf{T}_{2D}^{k-1,k}$ obtained in {\em step 2}\;
       \underline{\em Step 4}: For the new vertex, assess the
       registration quality $\varepsilon(V_e)$
       using~\eqref{eq:valVertex} and the images superimposed in {\em step 3}\;
     }

     \vspace*{.6mm}
     Store the six parameters of the vertex having the largest
     $\varepsilon(V_e)$ value in $D^{k-1,k}$, $\theta_1^{k-1,k}$,
     $\theta_2^{k-1,k}$, and $\theta_3^{k-1,k}$\;
     \vspace*{0mm}
     Fill matrix $\mathbf{T}_{3D}^{k-1,k}$ with the rigid
     transformation parameters computed from $D^{k-1,k}$,
     $\theta_1^{k-1,k}$, $\theta_2^{k-1,k}$, and $\theta_3^{k-1,k}$:
     see~\eqref{eq:matriceTrigideb};

     \vspace*{0mm}
     Fill matrix $\mathbf{T}_{2D}^{k-1,k}$ with the homography
     parameters corresponding to $\mathbf{T}_{3D}^{k-1,k}$.
\end{algorithm*}

\subsubsection{Similarity measure between
$I_{k-1}$ and $\mathbf{T}_{2D}^{k-1,k}(I_{k})$  \label{sec:similarityMeasure}}
The homography $\mathbf{T}_{2D}^{k-1,k}$ determined with
\eqref{eq:Lien2D3DMatricielFinal} is now used to superimpose image
$I_k$ on $I_{k-1}$. The hue being constant in bladder images
\cite{Miranda2008}, the pixel values of $I_k$ and $I_{k-1}$ are
converted into grey-levels for the registration step. As shown
in~\cite{Hernandez:2010}, homologous features cannot be systematically
extracted from the images due to blur, weak textures or lacking
textures in some image parts. Therefore, iconic data are directly used
to measure the similarity of the overlapping parts of $I_k$ and
$I_{k-1}$. The mutual information, denoted by ${\cal
  MI}(I_{k-1},\mathbf{T}^{k-1,k}_{2D}(I_{k}))$, is a robust similarity
measure \cite{Pluim:2003} which was successfully used for 2D bladder
image mosaicing~\cite{Miranda2008}. This statistical measure is
notably robust since it is computed with the whole pixels of the
common regions of the superimposed images.  The mutual information is
defined from the grey-level entropies ${\cal H}(I_{k-1})$ and ${\cal
  H}(\mathbf{T}^{k-1,k}_{2D}(I_{k}))$ and the joint grey-level entropy
${\cal H}_J(I_{k-1},\mathbf{T}^{k-1,k}_{2D}(I_{k}))$ of source image
$I_{k}$ and target image $I_{k-1}$:
\begin{eqnarray}
{\cal MI}\left(I_{k-1},\mathbf{T}^{k-1,k}_{2D}(I_{k})\right) =
{\cal H}(I_{k-1}) + \nonumber \qquad \qquad \\
\qquad \qquad {\cal H}(\mathbf{T}^{k-1,k}_{2D}(I_{k}))
- {\cal H}_J(I_{k-1},\mathbf{T}^{k-1,k}_{2D}(I_{k}))
\label{eq:MutualInformation}
\end{eqnarray}
with
\begin{align*}
\left\{\begin{array}{l}
{\cal H}(I_{k-1}) = -{\displaystyle\sum_{z_{k-1}} p(z_{k-1})\log p(z_{k-1})}\\
{\cal H}(\mathbf{T}^{k-1,k}_{2D}(I_{k})) = -{\displaystyle\sum_{
z_k} p(z_k)\log p(z_k)}  \\
{\cal H}_J(I_{k-1},\mathbf{T}^{k-1,k}_{2D}(I_{k})) = \\ \qquad \quad
-{\displaystyle\sum_{z_{k-1}}\sum_{z_k}
   p_J(z_{k-1},z_k)\log p_J(z_{k-1},z_k)}.
\end{array} \right.
\end{align*}
The grey-levels $z_{k-1}$ and $z_k$ range in $[0, 255]$ because images
are single-byte encoded. $p(z_{k-1})$ and $p(z_k)$ are the probability
density functions of the overlapping parts of images $I_{k-1}$ and
$\mathbf{T}_{2D}^{k-1,k}(I_{k})$ whereas $p_J(z_{k-1},z_k)$ is a joint
probability density function. The mutual information is maximal when
both images are registered, \emph{i.e.}, when $I_{k-1}$ is
statistically correlated with $\mathbf{T}_{2D}^{k-1,k}(I_{k})$.

\subsubsection{Registration algorithm\label{sec:simplex}}
The mutual information is maximized with respect to the rotation
angles $\theta_1^{k-1,k}$, $\theta_2^{k-1,k}$, $\theta_3^{k-1,k}$ and
the translation parameter $D^{k-1,k}$ appearing
in~\eqref{eq:matriceTrigideb}. The optimization task is done using the
simplex algorithm \cite{Nelder:1965} since the cost
function~\eqref{eq:MutualInformation} is non-differentiable. Because
there are six unknowns, the simplex is composed of seven vertices
$V_e$ (with $e \in \{1,\ldots,7\}$) lying in $\mathbb{R}^6$. The
similarity measure
\begin{align}
\varepsilon(V_e) = {\cal MI}(I_{k-1},\mathbf{T}^{k-1,k}_{2D}(I_{k}))
\label{eq:valVertex}
\end{align}
is computed for each vertex $V_e$, where the $\mathbf{T}_{2D}^{k-1,k}$
matrix deduces from $V_e$ as detailed in
\S~\ref{sec:matrixrelationship}. At each iteration, the vertex of
smallest $\varepsilon(V_e)$ value is replaced by a new vertex whose
location in the parameter space is computed with the simplex symmetry
rules. The angles and translations associated to the new vertex are
used to determine the rigid transformation matrix, compute the related
homography matrix, and then measure the image registration quality
using~\eqref{eq:valVertex}. By repeating this process, the simplex
vertices gradually converge towards the maximizer of
\eqref{eq:MutualInformation}. The optimization principle is detailed
in Algorithm \ref{alg:transformationDetermination}. The initial
simplex size is experimentally set relative to the 25 images/s
acquisition rate and the low cystoscope speed, leading to
sub-millimetric translations and to rotations smaller than $1^{\circ}$
between two acquisitions. The universal initial simplex is built with
0.3 mm translations and $0.4^{\circ}$ rotation values:
\begin{align}
%
\begin{array}{lclccccr}
V_1 & = & [\,0 & 0 & 0 & 0 &0 &0\,] \\
V_2 & = & [\,0.3 & 0 & 0 & 0 &0 &0\,] \\
V_3 & = & [\,0 & 0.3 & 0 & 0 &0 &0\,] \\
V_4 & = & [\,0 & 0 & 0.3 & 0 &0 &0\,] \\
V_5 & = & [\,0 & 0 & 0 & 0.4 &0 &0\,] \\
V_6 & = & [\,0 & 0 & 0 & 0 &0.4 &0\,] \\
V_7 & = & [\,0 & 0 & 0 & 0 &0 &0.4\,]
\end{array}
\label{eq:initialSimplex}
\end{align}
Each vertex gathers the three translation parameters (first three
entries in $D$), and then $\theta_1$, $\theta_2$ and $\theta_3$. The
iterative process of Algorithm~\ref{alg:transformationDetermination}
is stopped when the mean displacement of the corners of image $I_k$
(displaced in the $I_{k-1}$ domain) becomes smaller than 0.1 pixel.
When the optimization task is completed, the geometrical links between
consecutive coordinate systems $\{c_{k}$\} and $\{c_{k-1}$\} and
consecutive images ($I_k$, $I_{k-1}$) are fully defined by
$\mathbf{T}^{k-1,k}_{3D}$ and $\mathbf{T}^{k-1,k}_{2D}$, respectively.
%
%
%
%
\begin{figure*}[!t]
\centering
\begin{tabular}{c}
\includegraphics[width=0.8\textwidth]{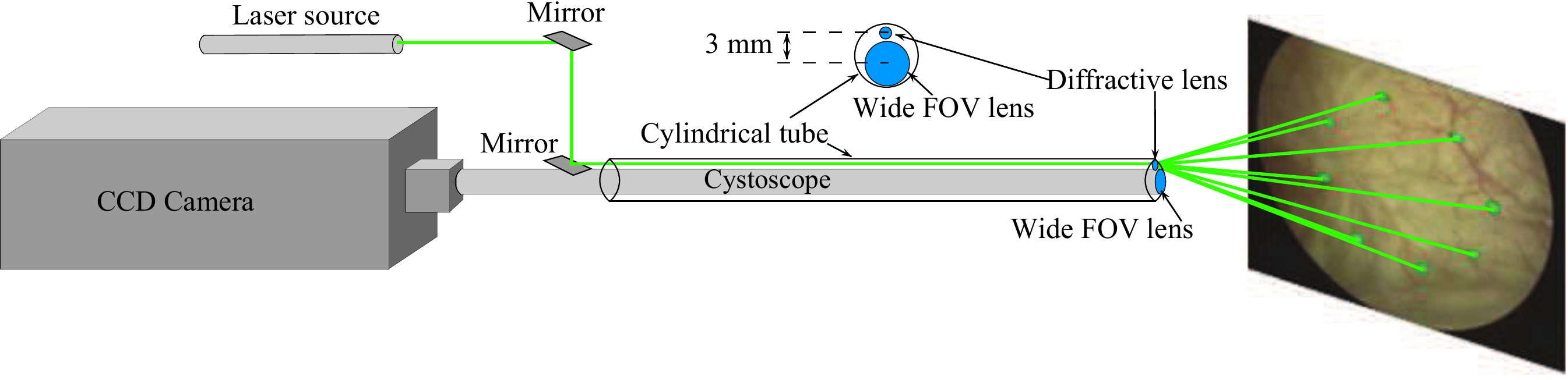}\\
(a)\\
\includegraphics[width=0.8\textwidth]{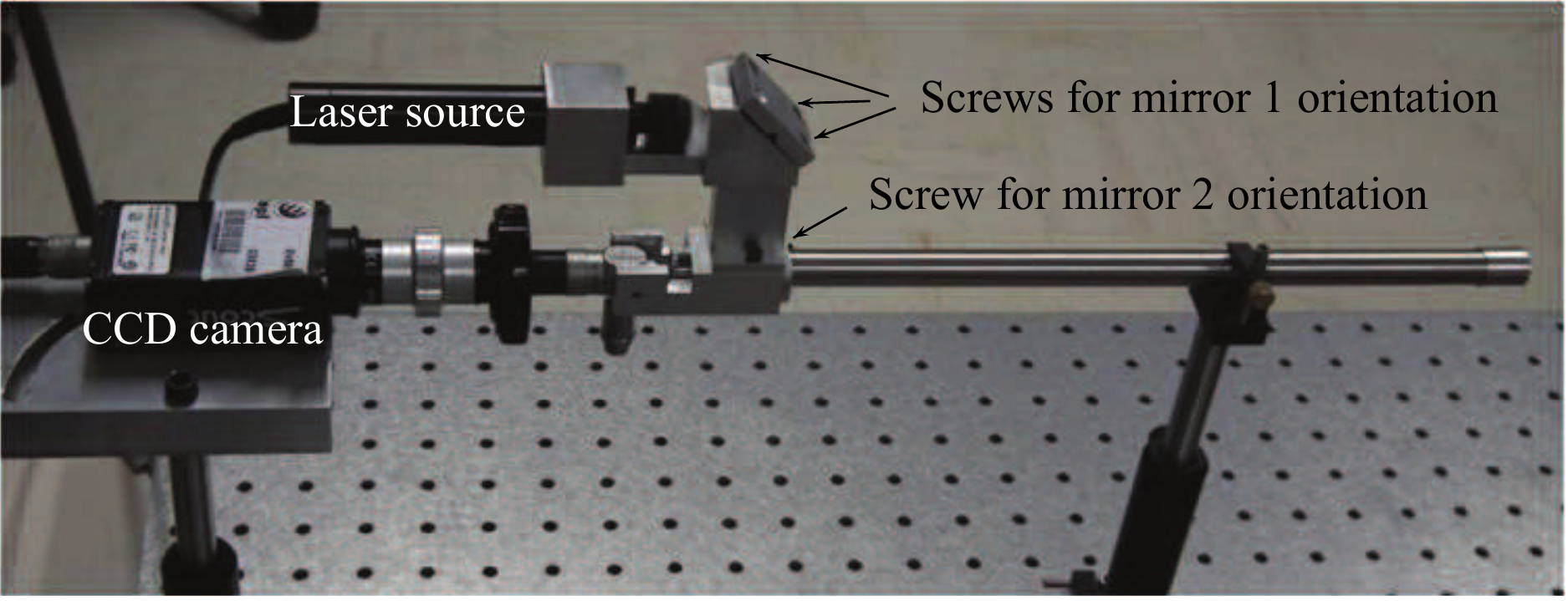}\\
 (b)
 \end{tabular}
 \caption{Active vision prototype. (a) Principle of the
   structured-light prototype. The trajectory of the beam emitted by
   the laser source is precisely controlled by two mirrors. The
   adjustable mirror orientations ensure that the laser beam falls on
   the center of the diffractive lens. This lens projects eight laser
   rays onto the scene. The laser wavelength and power were set to 532
   nm and 10 mW. The green color allows us to have a good contrast
   between the image textures and the green laser points. The 10 mW
   power was actually much too strong, therefore optical attenuators
   were used. The triangulation principle is sketched in
   Fig.~\ref{fig:2} and described in
   Section~\ref{sec:AcquisitionData}. The prototype (camera and
   projector) was calibrated using the generic method described in
   \cite{Ben-Hamadou:2013}.
   (b) Prototype corresponding to the sketch in (a). Mirrors 1 and 2
   are fixed onto planar metal supports. The orientation of the planar
   supports is precisely set using three screws. The planar support
   and two screws are not visible for mirror 2.}
\label{fig:3}
\end{figure*}
\subsection{Data mosaicing\label{sec:mosaicing}}
The 3D laser points from all viewpoints $k$ have to be expressed in a
common coordinate system. Choosing the camera coordinate system of the
first acquisition, the points of viewpoint $k$ are displaced from
$\{c_{k}$\} to $\{c_{1}$\} using the ``global'' (g) transformation
matrix:
\begin{equation}
\mathbf{T}_{3D}^{k,g} = \prod_{j=2}^{k} \mathbf{T}_{3D}^{j-1,j} = \mathbf{T}_{3D}^{1,2}
\times \mathbf{T}_{3D}^{2,3}\times \hdots \times \mathbf{T}_{3D}^{k-1,k}.
\label{eq:discreteSurface}
\end{equation}
The whole set of points in $\{c_{1}$\} is then exploited to compute a
3D mesh with triangular faces representing the bladder
surface~\cite{Kazhdan:2006}. Finally, the textures of images $I_k$ can
be projected onto the 3D mesh.
Specifically, the color of each triangular face in the 3D mesh is
obtained by copying the color of the pixel in the first image of the
video in which the triangular face can be seen. This simple choice
could be improved using more involved strategies in order to reduce
possible texture discontinuities in the 3D mosaic or to smooth the 3D
surface by thin-plate interpolation. See \emph{e.g.},
\cite{Weibel2012b} in the 2D case.

Similar to the 3D case, a 2D mosaic is built by taking the coordinate
system of the first image as reference and using the global 2D
matrices:
\begin{equation}
\mathbf{T}_{2D}^{k,g} = \prod_{j=2}^{k} \mathbf{T}_{2D}^{j-1,j} = \mathbf{T}_{2D}^{1,2}
\times \mathbf{T}_{2D}^{2,3}\times \hdots \times \mathbf{T}_{2D}^{k-1,k}.
\label{eq:2DMosaic}
\end{equation}

It has to be noticed that the green laser points cover a small surface
in the images and are easy to segment since a simple thresholding of
the H (hue) component in the HSV color space robustly separates the
green laser points from the reddish or orange bladder epithelium
pixels (see \cite{daul:2000} for a detailed description of the HSV
color space family).  Since images $I_k$ and $I_{k-1}$ include large
common scene parts, the missing bladder textures in $I_{k-1}$ can be
replaced in the mosaic by the homologous texture pixels of $I_k$.  The
correspondence between pixel textures hidden by laser dots in image
$I_{k-1}$ and the corresponding visible texture pixels in image $I_k$
is given by homography $T_{2D}^{k-1,k}$. This correspondence is used
to build 2D and 3D mosaics without laser dot pixels.
%
%
%
%
%
\begin{figure*}[!t]
\centering
\begin{tabular}{ccc}
\includegraphics[height=4.3cm]{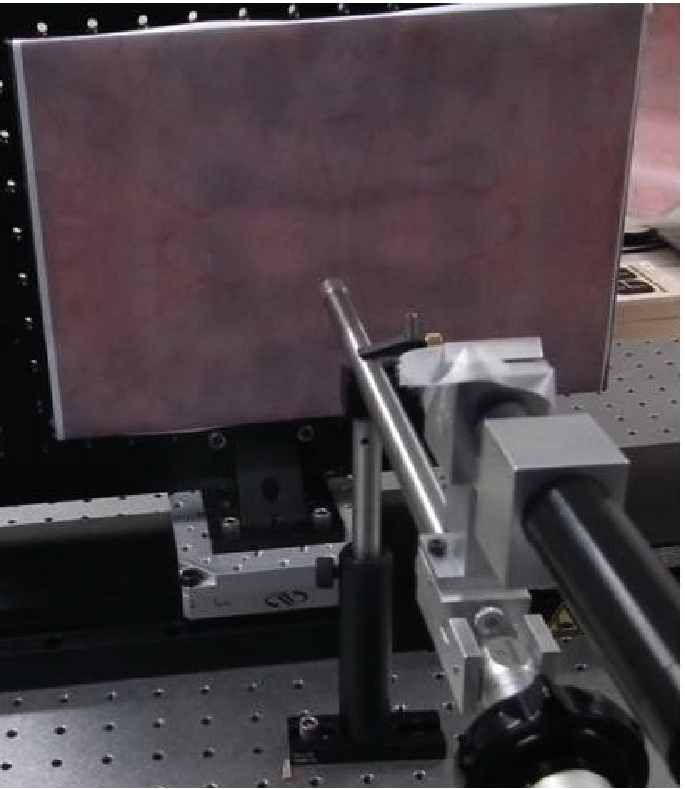}&
\includegraphics[height=4.3cm]{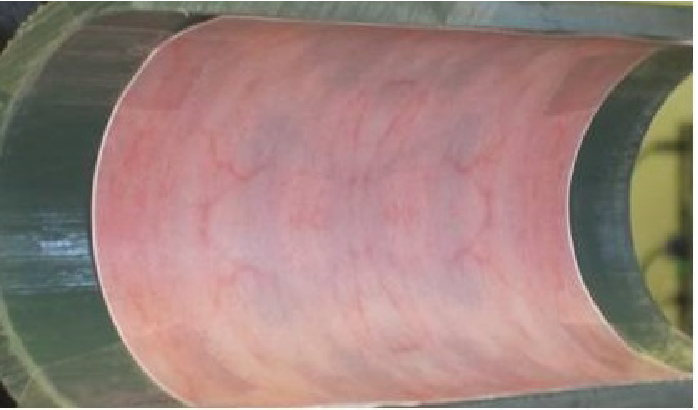}&
\includegraphics[height=4.3cm]{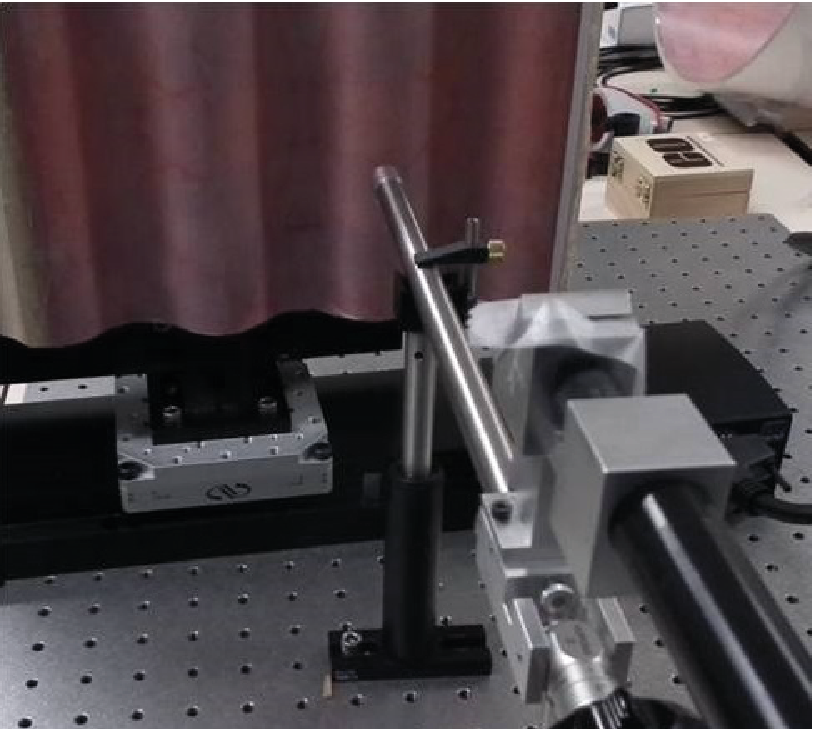}\\
(a)&(b)&(c)
\end{tabular}
\caption{Three real phantoms. The planar phantom (a) is used to assess
  the accuracy of the 3D registration algorithm. The half cylinder~(b)
  and wave~(c) phantoms allow us to evaluate the ability of the
  registration method to construct convex and nonconvex surface parts,
  respectively.}
\label{fig:4}
\end{figure*}

\section{Quantitative results for prototype data \label{sec:ProtoResultsDiscussion}}
The goal of this section is to validate on real data that the
mosaicing approach of Section~\ref{sec:3Dmosaicing} indeed leads to
coherent surface construction results. To do so, an active vision
prototype is first proposed. Then, 3D mosaicing results are presented
for data acquired with the prototype using three real phantoms, on
which pig bladder texture images were stuck. This allows for
a first quantitative assessment of the mosaicing accuracy based on ground
truths. As mentioned before, our prototype cannot be used in clinical
situations yet. It was rather designed for validation of the
structured-light approach in combination with a cystoscope and to show
the feasibility of 3D cartography. The applicability to clinical
situations will be further discussed in
Sections~\ref{sec:SimulatedDataResultsDiscussion} and
\ref{sec:Conclusion} together with complementary tests on realistic
simulated data.

\subsection{Prototype and test phantom description \label{sec:ProtoPhantomDescription}}
Our active vision prototype is illustrated in Fig.~\ref{fig:3}. The
acquisition channel consists of a Karl Storz cystoscope, and the image
sequences are acquired with a Basler scout model camera. The image
resolution is 768 $\times$ 576 pixels. The diffractive lens is fixed
nearby the wide FOV lens of the cystoscope distal tip. The principle
of the laser ray projection is sketched in Fig.~\ref{fig:3}(a). This
prototype uses the triangulation principle described in
Section~\ref{sec:AcquisitionData} and has a geometry equivalent to
that of Fig.~\ref{fig:2}(a). The baseline (distance between the camera
and projector optical centers, see Fig.~\ref{fig:3}(a)) available for
triangulation is about 3 mm.
The laser projector wavelength is equal to 520 nm, and the 10 mW power
should be adapted for further clinical deployment.

High resolution images of an incised and flattened pig bladder were
acquired to simulate human bladder textures (see
Fig.~\ref{fig:4}(b)). The images were printed on paper sheets and
stuck on surfaces with various shapes. First, the planar surface of
Fig.~\ref{fig:4}(a) is used to test the 3D mosaicing algorithm. Here,
the underlying assumption in the proposed algorithm is perfectly
fulfilled: the surfaces viewed in consecutive images are planar. Then,
the half cylinder of Fig.~\ref{fig:4}(b) is used to simulate a convex
surface. The half cylinder has a radius of 35 mm and allows for
simulating a plausible bladder curvature in the direction orthogonal
to the main cylinder axis (a non-warped bladder can be approximately
thought of as an ellipsoid with a major axis from 80 to 100 mm and
minor axes from 30 to 40 mm). The bladder is in contact with other
organs is often warped. In this case, the surface includes concave
parts due to troughs in the organ wall. As shown in
Fig.~\ref{fig:4}(c), a wave phantom was built to simulate warped
bladder parts with troughs between waves. The length of the wave (a
period of the phantom) is 40 mm and the wave depth (from the trough to
the crest of the wave) is 20 mm. This phantom simulates extreme
situations since consecutive troughs of large magnitude never arise in
practice.
All phantoms are positioned and displaced relative to the endoscope
prototype using a micrometric positioning table to generate ground
truth 3D transformations.
%
%
%
%
%
%
\begin{figure*}[!t]
\centering
%
\includegraphics[width=0.7\linewidth]{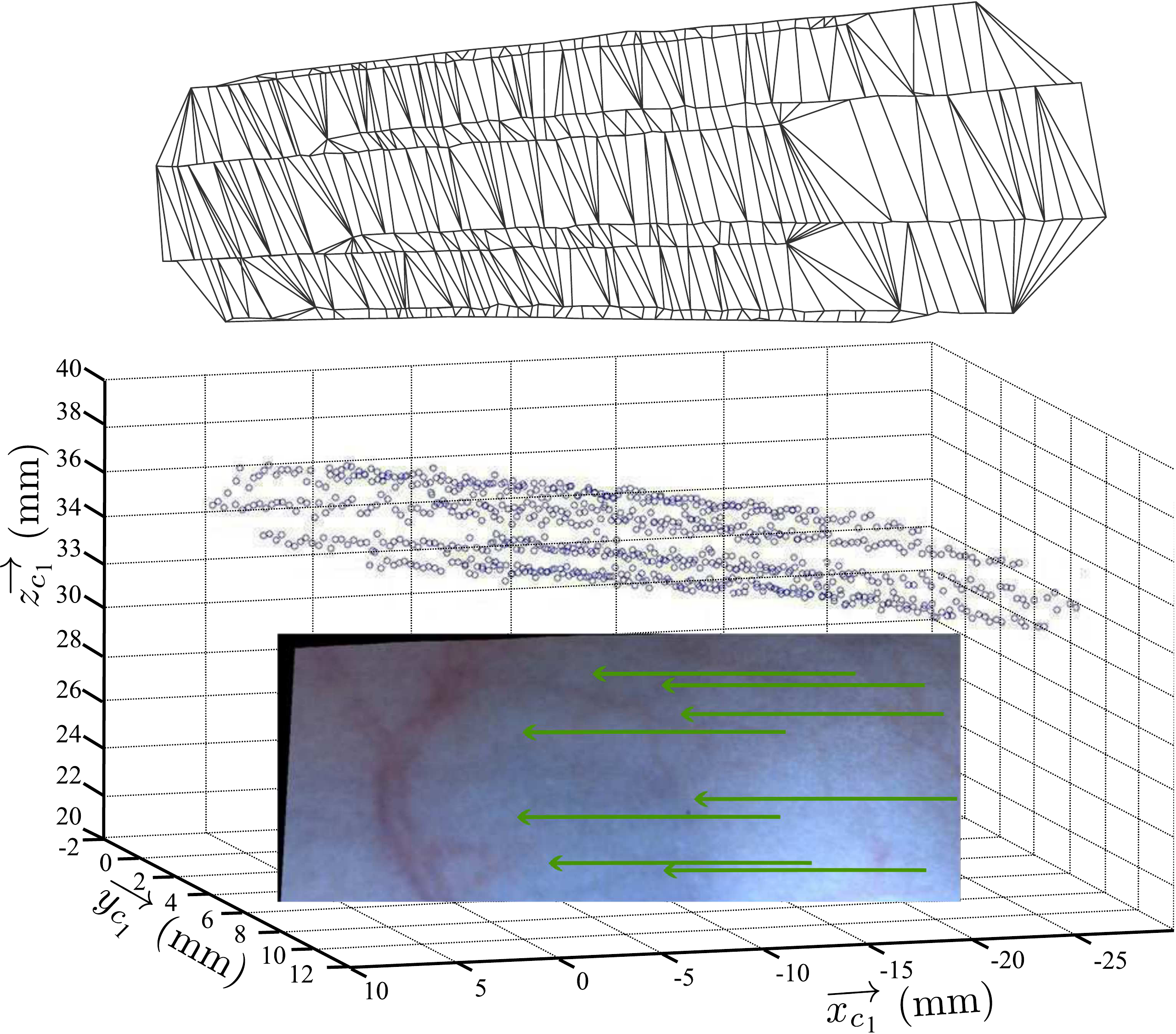}
%
\caption{Plane construction results. Bottom: 2D mosaic. The green
  arrows refer to the 8 laser point displacements (constant
  translation, no rotation) throughout image acquisition. Center: 3D
  points placed in the common coordinate system $(O_{c_1},
  \protect\overrightarrow{x_{c_1}}, \protect\overrightarrow{y_{c_1}},
  \protect\overrightarrow{z_{c_1}})$. Top: related mesh with
  triangular faces on which the 2D mosaic image can be projected.\label{fig:5}}
\end{figure*}
\subsection{Point reconstruction accuracy and surface construction results
\label{sec:ResultsRegisMosaics}}
Once the prototype has been calibrated with the method described
in~\cite{Ben-Hamadou:2013}, 3D point reconstruction is performed in
the local camera coordinate system $\{c_k\}$, and $M=8$ $P_{3D}^{i,k}$
points are then available per viewpoint $k$.

In order to evaluate the laser point reconstruction accuracy, a first
data acquisition was carried out using the wave phantom. The
cystoscope axis was held orthogonal to the translation plane and
$K=200$ images were acquired with 1 mm translation steps. The range of
$z_{c_k}$-values that were found (in between 20 and 40 mm) for the
$P_{3D}^{i,k}$ points matches the wave height, equal to 20 mm. It was
further verified that for prototype distal tip-to-surface distances
ranging from some millimeters to 5 centimeters, the average
reconstruction error on the $z_{c_k}$ coordinate (depth) is close to
0.5 mm and remains submillimetric for all points. When normalizing
this 3D reconstruction error by the tip-to-surface distance, the
errors are always lower than $3~\%$. This accuracy is by far
sufficient for bladder surface mosaicing since the constructed surface
shape must only be representative of the organ wall (a precise
dimensional measure does not make sense in urology).

In the following surface construction tests and without loss of
generality, the phantoms were displaced in front of the fixed
prototype. This is in contrast with the setup of
Section~\ref{sec:3Dmosaicing}, where the instrument is moving inside
the organ. However, it is easy to see that the proposed mosaicing
algorithm remains valid and unchanged in both settings whatever the
moving object (surface or instrument). The simultaneous 2D and 3D data
mosaicing results are illustrated in Fig.~\ref{fig:5}
for the planar phantom. Here, 100 images were acquired for 99 constant
translations of 0.3 mm (with the three components of $D^{k-1,k}$
different from 0) and without 3D rotations. Thus, the magnitude of 3D
displacements is equal to:
\begin{align}
\|D^{k-1,k}\|=0.3 \label{eq:criterionTranslations}
\end{align}
where $\|\,.\,\|$ stands for the Euclidean norm.
%

%
%
%
\begin{figure*}
\centering
\psfig{figure=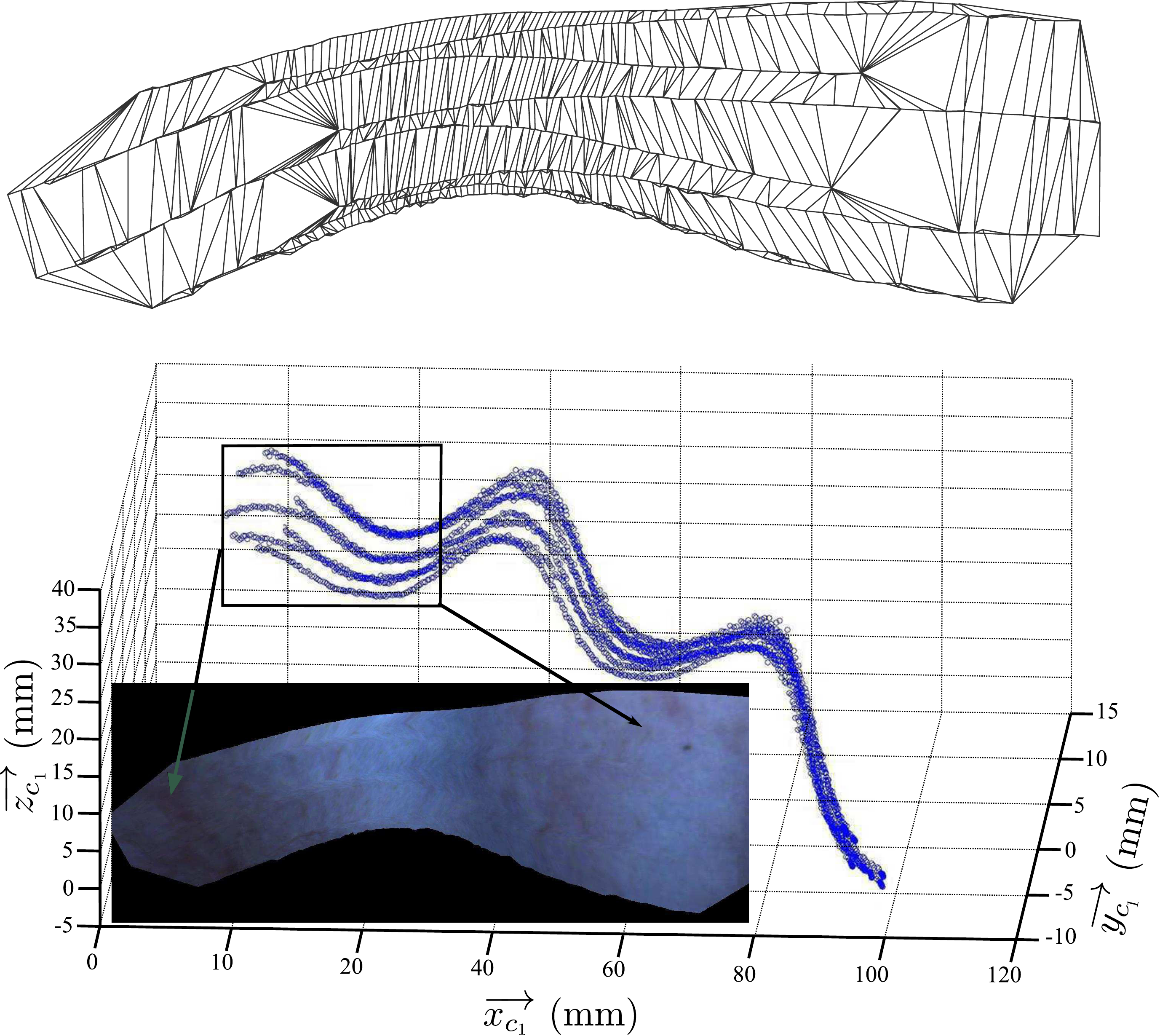, width=0.8\linewidth}
  \caption{Wave construction results. Center: 3D points placed in the
    common camera coordinate system $(O_{c_1},
    \protect\overrightarrow{x_{c_1}},
    \protect\overrightarrow{y_{c_1}},
    \protect\overrightarrow{z_{c_1}})$ and representing 2.5 wave
    periods in the phantom of Fig.~\ref{fig:4}(c). Top: mesh surface
    reconstructed from the first 100 acquisitions (zoom in
    rectangle). Bottom: textured surface corresponding to the first
    100 acquisitions. The cystoscope prototype is roughly oriented
    along the $\protect\overrightarrow{z_{c_1}}$ axis. The local
    phantom surface corresponding to the left edge of the zoom in
    rectangle is the most distant from the camera in contrast with the
    local surface related to the right edge.\label{fig:6}}
\end{figure*}

The average translation magnitude and standard deviation computed with
the registration algorithm for the 99 consecutive image pairs are 0.27
$\pm$ 0.015 mm, whereas the angles $\theta_1^{k-1,k}$,
$\theta_2^{k-1,k}$ and $\theta_3^{k-1,k}$ indeed tend towards
$0^{\circ}$. So, the average magnitude of displacement is slightly
underestimated with respect to the ground truth (0.3 mm), whereas the
$0.015$ mm standard deviation value is very low. These results are a
first indication that the nature (translation) of the displacement can
be successfully determined using the registration algorithm.
Fig.~\ref{fig:5} confirms that accurate results are obtained for the
planar phantom. The 2D mosaic built from the estimated
$\mathbf{T}_{2D}^{k-1,k}$ homographies is visually coherent,
\emph{i.e.}, without texture discontinuities between images. In the
textured image of Fig.~\ref{fig:5}, the green arrows represent the
trajectory of the eight laser points throughout the video-image
acquisition. As could be expected, these trajectories are affine
because the camera movement is a pure translation. The 3D points,
placed in the common coordinate system $\{c_{1}$\}, all lie on a
plane. The constructed mesh with triangular faces is displayed in the
top image of Fig.~\ref{fig:5}.
%
%
%
%
%
%
\begin{figure*}
\begin{tabular}{cc}
\setlength{\tabcolsep}{0.05cm}
\begin{tabular}{c}\includegraphics[width=10cm]{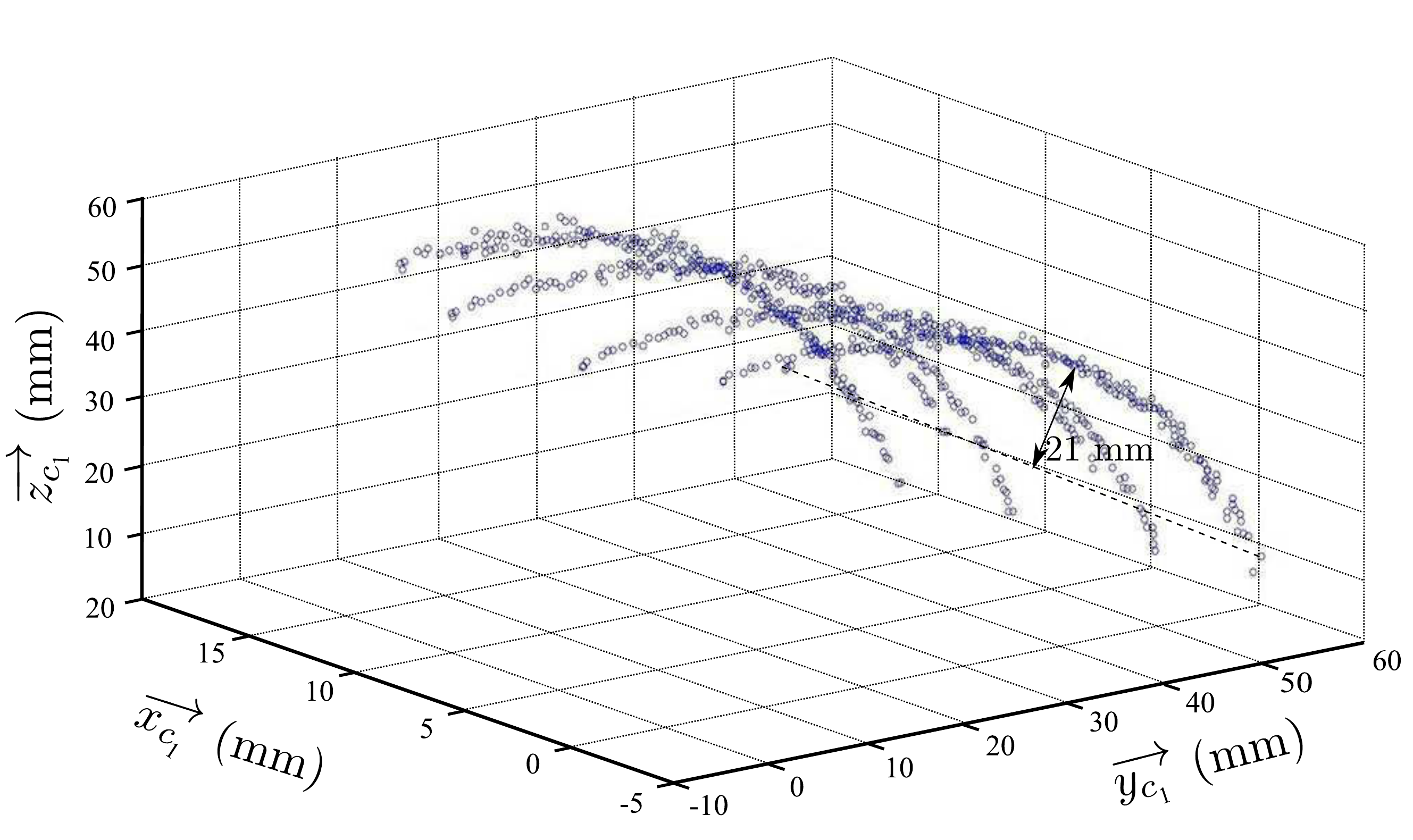}\end{tabular}&
\begin{tabular}{c}\includegraphics[width=6cm]{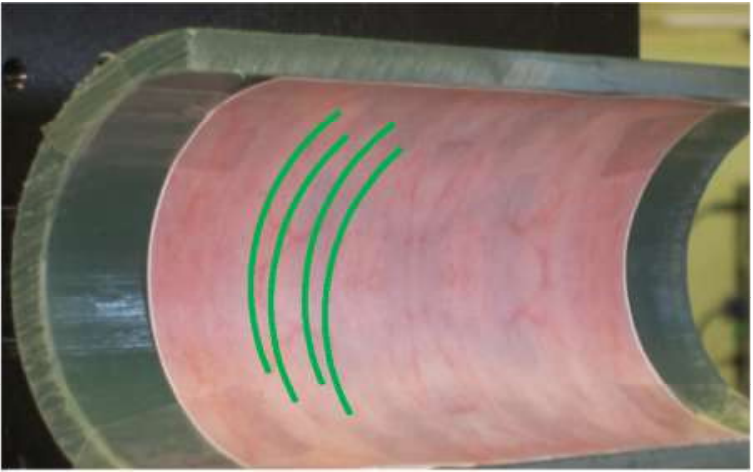}\end{tabular}\\
(a)&(b)
\end{tabular}
\caption{Half cylinder reconstruction results. (a)~3D points placed in
  the global coordinate system $\{c_{1}$\}. Their locations are
  consistent with the trajectories sketched in green on the half
  cylinder phantom~(b). Each green curve represents the trajectory of
  two very close laser points.}
\label{fig:7}
\end{figure*}

In the following surface construction tests, we aim to address a case where the
geometrical link between the surfaces in the FOV of two images is not
exactly an homography (working assumption of the 3D registration
algorithm) since the local surfaces are not planar anymore. The data
of Fig.~\ref{fig:6} were acquired for the wave phantom with the same
displacements as before: acquisition of $K=360$ images for 359
constant translations of 0.3 mm and without any rotation. For this
phantom, the local surface orientation with respect to the cystoscope
optical axis strongly varies throughout the sequence. To quantify the
algorithm accuracy, we used again
criterion~\eqref{eq:criterionTranslations}. The average magnitude of
the 3D translations computed with the registration algorithm for
consecutive image pairs is $0.29 \pm 0.03$ mm, and
$\theta_1^{k-1,k}=\theta_2^{k-1,k}=\theta_3^{k-1,k} \approx
0^{\circ}$. The magnitude of translations (0.29 mm) is very close to
the true average displacement (0.3 mm) whereas the related standard
deviation (0.03 mm) remains very close to the ideal zero value. These
results show that the translation of the cystoscope prototype can be
successfully determined for non-planar surfaces and for varying angles
between the local surface and the cystoscope optical axis. The
variation of this angle combined with changes of phantom-to-camera
distances (see Fig.~\ref{fig:6}) impact the perspective
($a_{3,1}^{k-1,k}$ and $a_{3,2}^{k-1,k}$), shearing ($s_x^{k-1,k}$ and
$s_y^{k-1,k}$), and scale ($\alpha^{k-1,k}$) parameters,
see~\eqref{eq:matriceHomography}.

The blue points in Fig.~\ref{fig:6} represent again the 3D laser
points placed in the common coordinate system $(O_{c_1},
\overrightarrow{x_{c_1}}, \overrightarrow{y_{c_1}},
\overrightarrow{z_{c_1}})$. The 360 viewpoints correspond to 2.5 wave
periods, \emph{i.e.}, a length of 100 mm. Qualitatively, the global
surface indeed corresponds to waves although after 1.5 wave period,
the wave surface bends itself instead of remaining straight. However,
the results are satisfactory since a wave period represents an organ
pushing against the bladder (\emph{e.g.}, a trough), and two or three
consecutive troughs are an extreme situation never arising for
cystoscopic data.

In Fig.~\ref{fig:7}, the cystoscope prototype was approximately held
at a 30 mm distance from the half cylinder surface (of radius 190 mm)
during the acquisition of 130 images. The surface covered by the
cystoscope corresponds to a circular arc of aperture $130^\circ$. The
displacements between consecutive acquisitions $k-1$ and $k$ mainly
consist of $1^\circ$ rotations around $\overrightarrow{x_{c_1}}$
combined with small translations and rotations around
$\overrightarrow{y_{c_1}}$ and $\overrightarrow{z_{c_1}}$.
Fig.~\ref{fig:7} shows that the reconstructed points indeed lay on a
cylindrical surface. The depth of the cylinder part was used to
quantify the surface construction accuracy. Ideally, the dashed line in
Fig.~\ref{fig:7}(a) should circumscribe an arc of aperture
$130^\circ$. The real depth of the cylinder part corresponding to the
$130^\circ$ phantom aperture is 25 mm, whereas the depth computed from
the reconstructed data is 21 mm. This error does not alter the visual
quality of the surface representation: the phantom shape is indeed
cylindrical and no dimensional analysis is required.

\subsection{Overview of the results related to prototype
  data \label{sec:DiscussionProtoDataResults}}
The results obtained so far are a first validation that the prototype
displacement (\emph{i.e.}, the displacement of the camera coordinate
system in between two acquisitions) can be recovered using only the
video-image and 8 laser points per acquisition. Indeed, for both
phantom types, the deviation of the estimated 3D displacement from the
ground truth is small.

{\em - Local surface planarity.} Planarity is one of the main
differences between the plane and wave phantoms. Indeed, the planar
assumption (made for registering two images using an homography) is
not fulfilled anymore for the surface parts of the wave. We found that
the camera displacements were computed with similar accuracy for both
planar and non-planar surfaces. It was already proved in the frame of
2D bladder mosaicing that image registration with homographies is
accurate even for regions where the bladder is
non-planar~\cite{Weibel:2012a} or contains lesions such as
polyps~\cite{Hernandez:2010}. The results of
\S~\ref{sec:ResultsRegisMosaics} indicate that homographies linking 2D
images are appropriate as well for 3D mosaicing.

{\em - 3D point reconstruction accuracy.} Inaccurate 3D laser
  point reconstruction impacts the precision of 3D camera displacement
  estimation. The results obtained with real phantoms show that 3D
  point reconstruction is accurate enough to enable camera motion
  assessment and 3D mosaicing, even for active vision prototypes
  having small baselines (3 mm here).

{\em - Cystoscope axis orientation.} For tests with wave phantoms,
  the angle between the cystoscope optical axis and the normal to the
  local surface is varying: at extreme points (points of minimum or
  maximum depth) of the wave, the angles are small ($\approx
  0^\circ$), whereas for medium points, their value is maximum
  ($45^{\circ}$). The accuracy of the reconstructed camera
  displacement is not strongly impacted by the variation of angles.

  The tests with the plane and wave phantoms involved 3D translations,
  whereas the viewpoint changes arising in the half-cylinder
  experiment combined both rotations and translations. In the latter
  case, no ground truth displacements were available. However, the
  cylinder depth that was estimated for an $130^\circ$ aperture shows
  that the surface shape can be accurately recovered.

  In all tests, images were acquired for a distal tip-to-surface
  distance ranging from 10 to 40 mm and for surface areas of one to
  several square centimetres. These dimensions are consistent with
  real bladder image acquisitions in terms of acquisition distance
  variation and size of the area viewed in images. For this distance
  interval, the laser points were systematically visible in the
  images, and their position could then be reconstructed in the camera
  coordinate system.
%
%

%
%
%
\begin{figure}[!t]
%

\centerline{\includegraphics[width=0.8\linewidth]{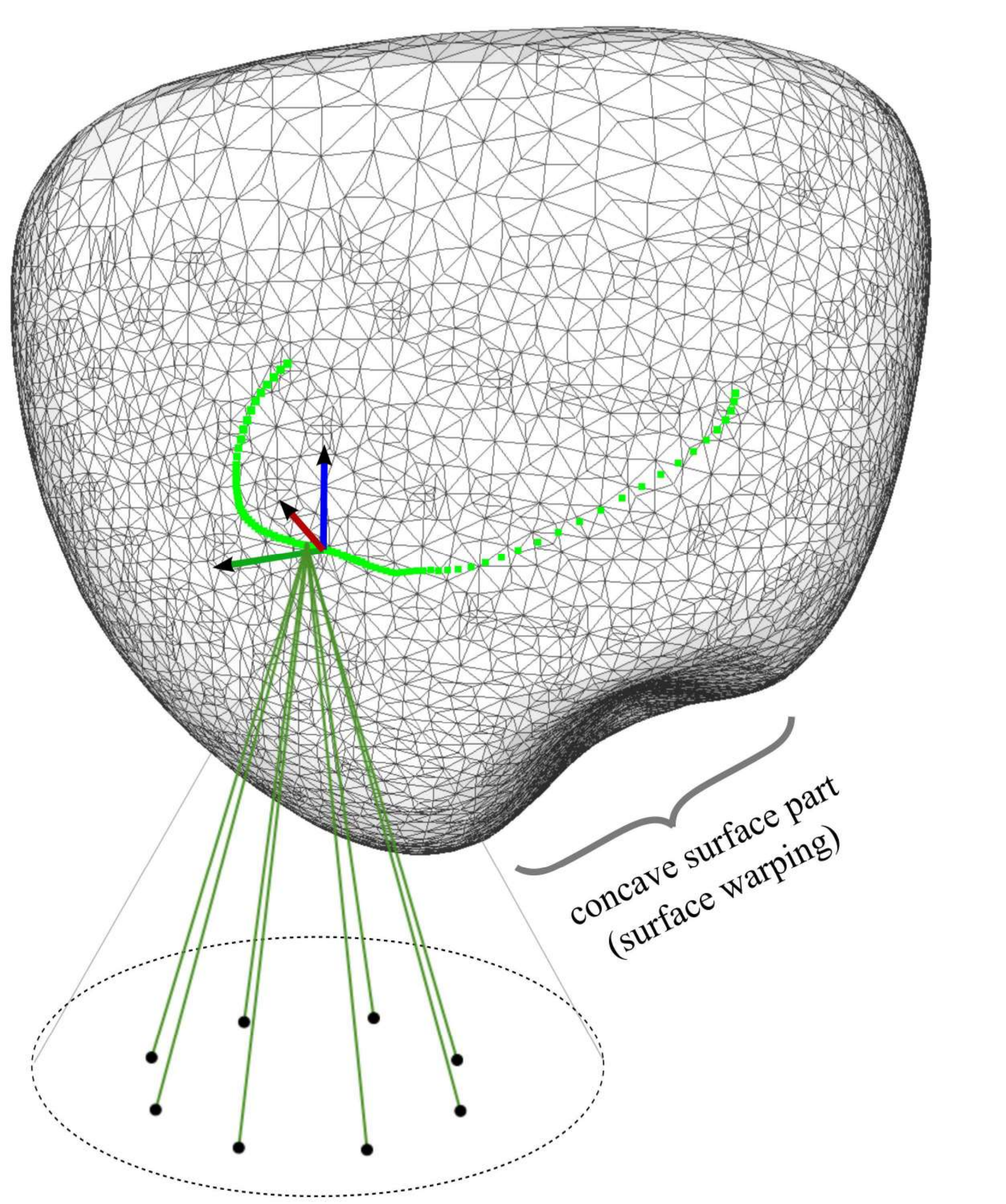}}
\caption{Simulated bladder volume and simulation of video sequence
  acquisition with known 3D endoscope displacements. The cone
  corresponds to the field of view of the camera for one viewpoint.
  The green dashed curve refers to the trajectory of the virtual
  cystoscope. The simulated video sequence then contains images
  related to both convex and nonconvex parts of the virtual bladder. }
\label{fig:8}
\end{figure}
%
%

%
%
%
\begin{figure*}[!t]
\centering
\begin{tabular}{ccc}
\includegraphics[width=0.3\linewidth]{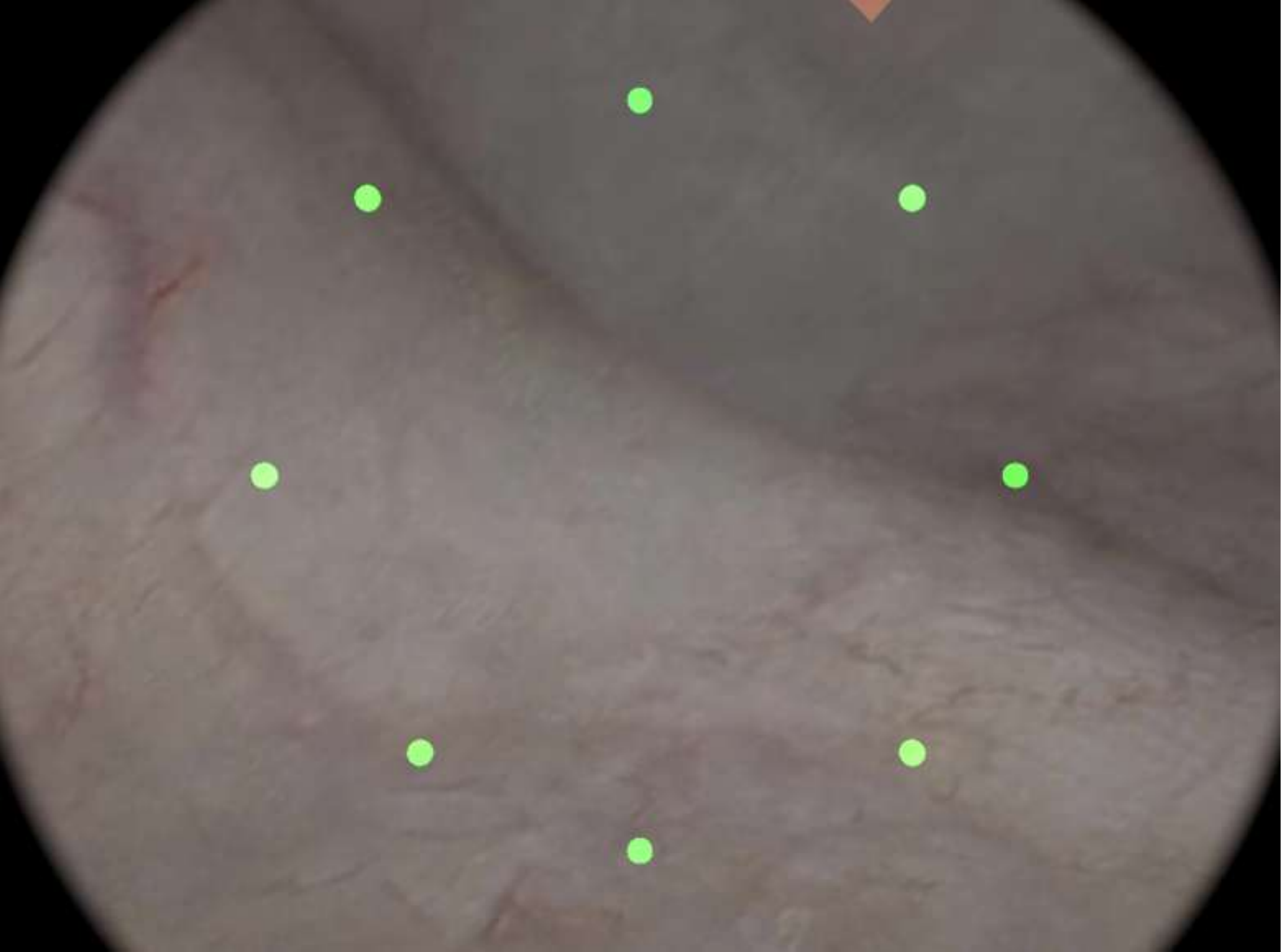} &
\includegraphics[width=0.3\linewidth]{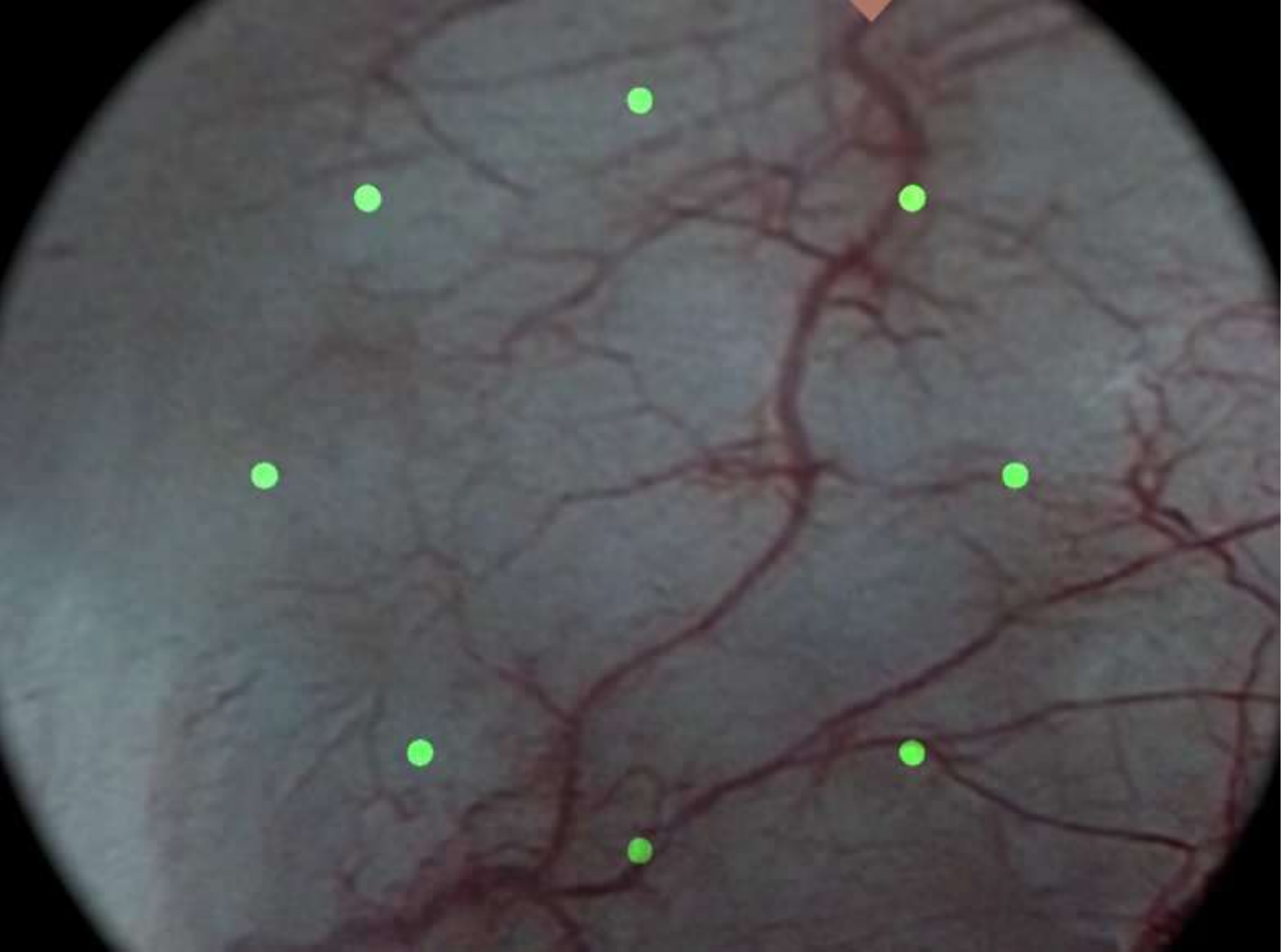} &
\includegraphics[width=0.3\linewidth]{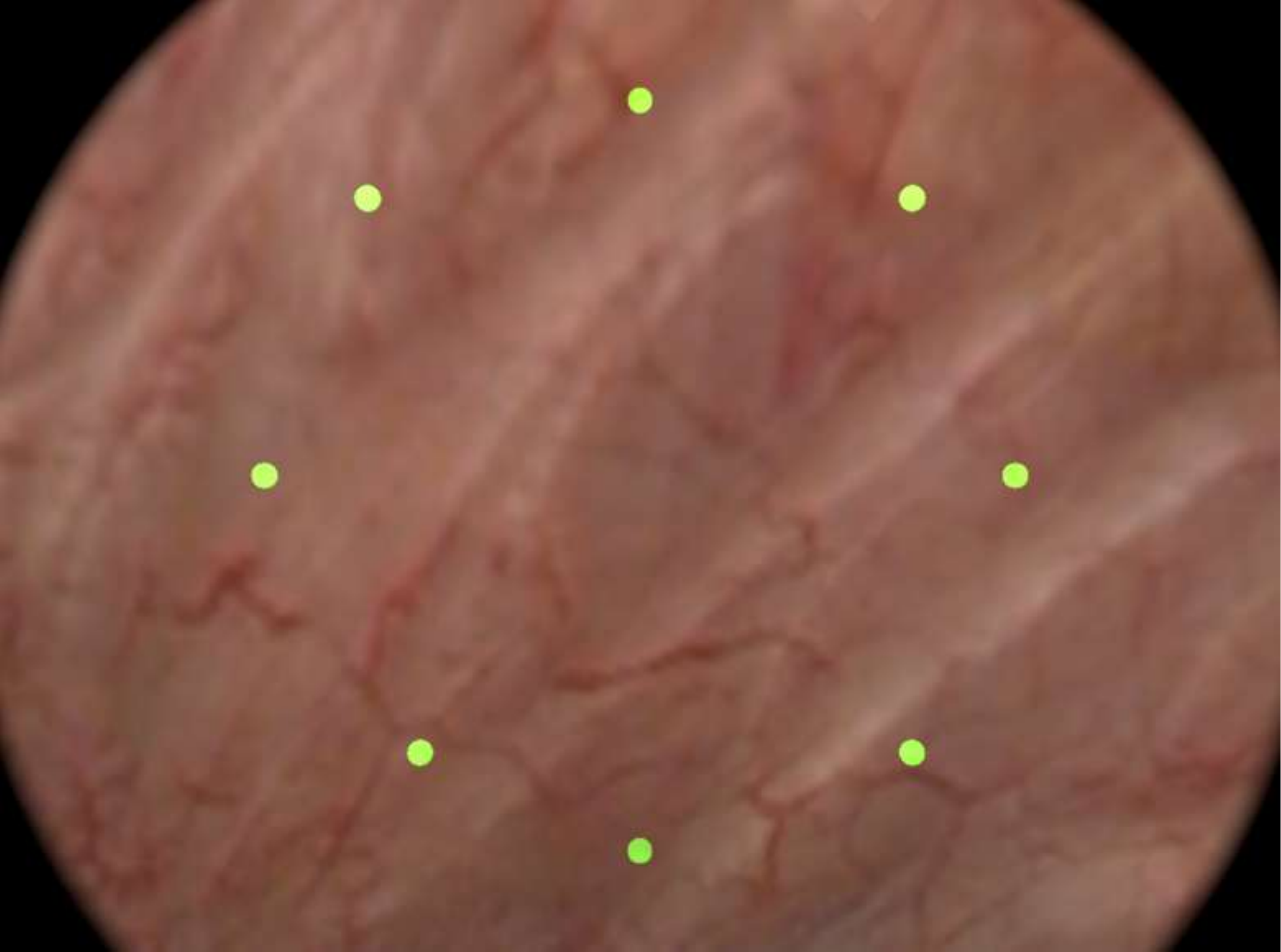} \\[.2cm]
(a)&(b)&(c)
\end{tabular}
\caption{Extraction of an image from each of the three simulated
  video-sequences. (a) Image with weakly contrasted texture.  (b)
  Image with highly contrasted vessels. (c) Image with blurry
  texture.}
\label{fig:9}
\end{figure*}

\section{Results for realistic simulated data \label{sec:SimulatedDataResultsDiscussion}}
The results presented so far are a first validation of the feasibility
of 3D bladder mosaicing. However, similar to the contributions of
Soper \emph{et al}~\cite{Soper2012} and Shevchenko \emph{et
  al}~\cite{Shevchenko:2012}, the method was not tested on patient
data (our active vision prototype was not built for patient data
acquisition at this stage). In this section, realistic bladder data
are simulated to perform complementary mosaicing tests. Not only the
bladder volume and texture are closer to human data, but also the
simulated endoscope trajectory can be controlled to reproduce all
types of cystoscope displacements.

\subsection{Simulation of textured bladder data}
To simulate a more realistic bladder shape, we considered a (convex)
oval volume whose width, height and depth are equal to 110, 100, and
70 mm, respectively (the standard capacity/volume of the bladder
ranges in between 400 and 600 ml). The ovoid was then locally deformed
to create a nonconvex surface part. This surface, illustrated in
Fig.~\ref{fig:8}, was defined in agreement with the Institut de
Canc\'erologie de Lorraine. Human bladder textures were extracted from
large FOV mosaics built using 2D mosaicing algorithms
\cite{Miranda2008,Weibel:2012a}. Three such mosaics were considered
(see Fig.~\ref{fig:1}) and mapped on the inner wall of the simulated
3D bladder surface.

Blender (a 3D graphics modeling and rendering
software~\cite{blender:2014}) was used to simulate the camera and the
laser projector while reproducing the geometrical configuration of our
prototype. Such 3D representation can be found in Fig.~\ref{fig:8}.
Both camera and laser projector parameters (calibrated
using~\cite{Ben-Hamadou:2013}) are handled. Moreover, Blender allows
us to simulate complex cystoscope trajectories combining rotations and
translations. Similar to clinical examinations, the distance between
the cystoscope distal tip and the surface is changing. The resulting
area seen in images varies from 1 to a few square
centimeters. Fig.~\ref{fig:9} displays a sample of the simulated
video-sequences obtained for each of the three numerical phantoms
using the 3D surface of Fig.~\ref{fig:8}. Each image sequence includes
120 images and covers $24$ cm$^2$ of the phantom surface. Importantly,
the texture variability is large from one video-sequence to another.
%
%
%
%
%
\begin{figure*}
\centering
{
\setlength{\tabcolsep}{-0.2cm}
\begin{tabular}{ccc}
\includegraphics[width=0.36\textwidth]{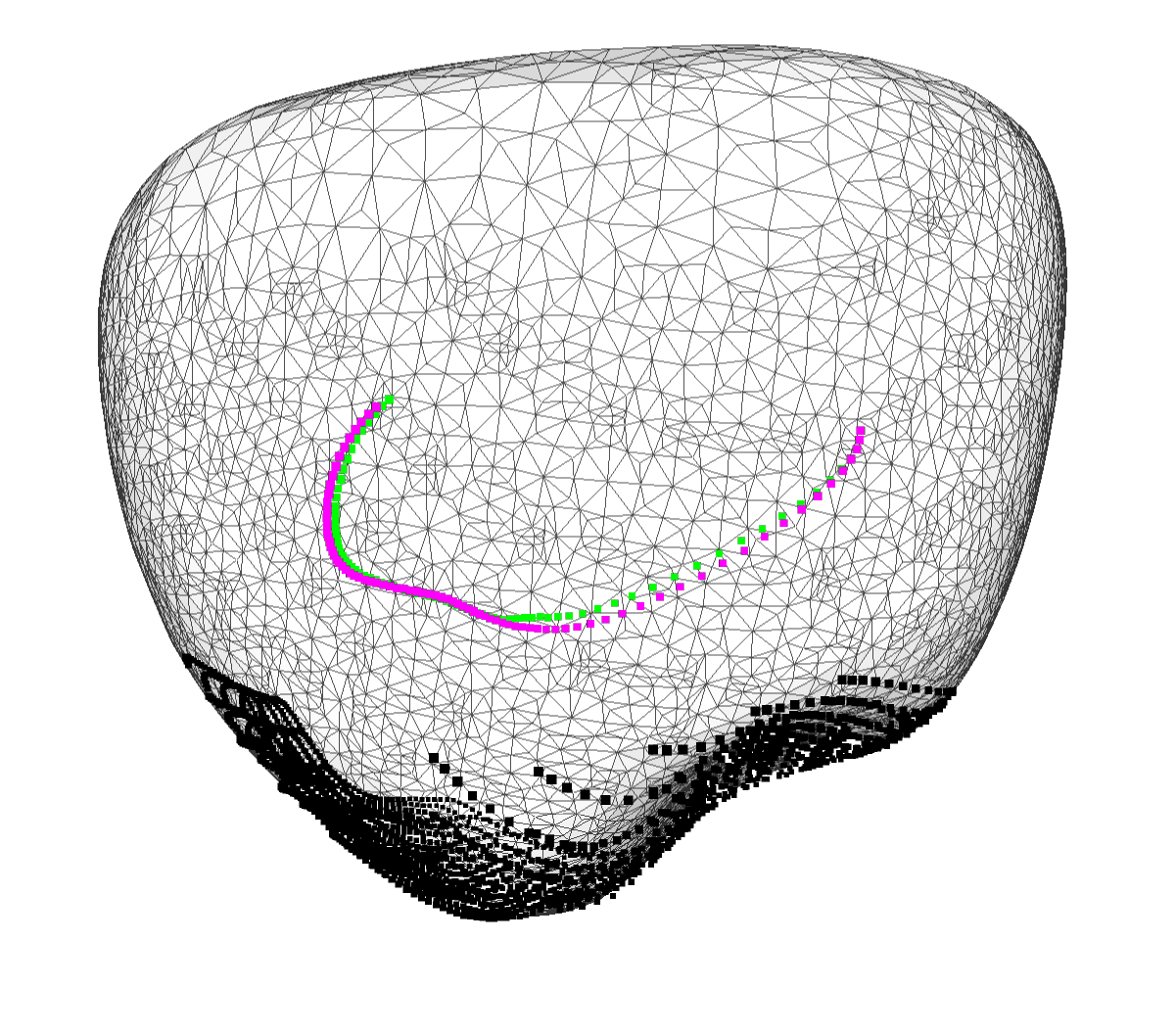} &
\includegraphics[width=0.36\textwidth]{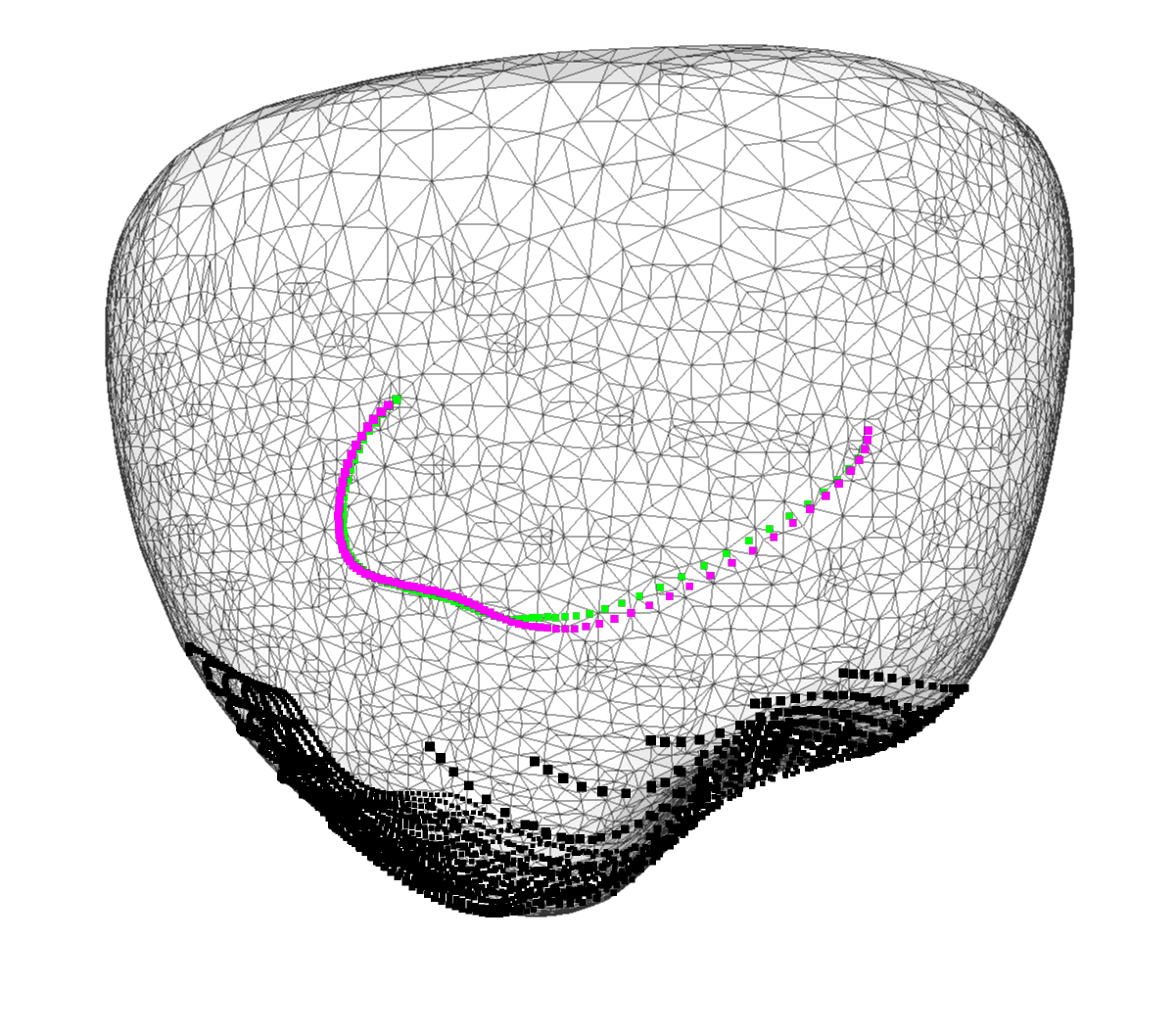} &
\includegraphics[width=0.36\textwidth]{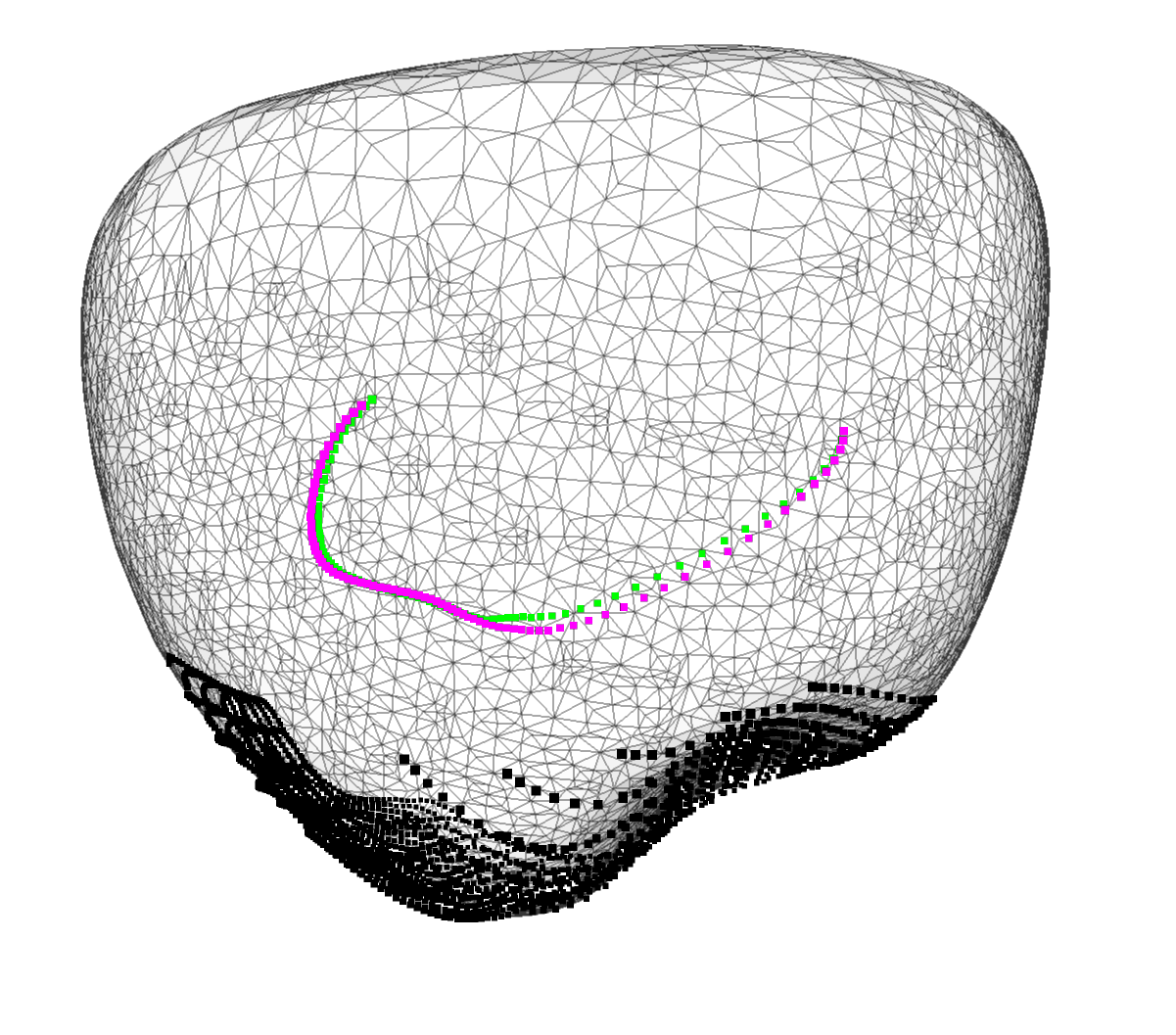} \\[-.4cm]
(a)&(b)&(c)\\[.4cm]
\includegraphics[width=0.36\textwidth]{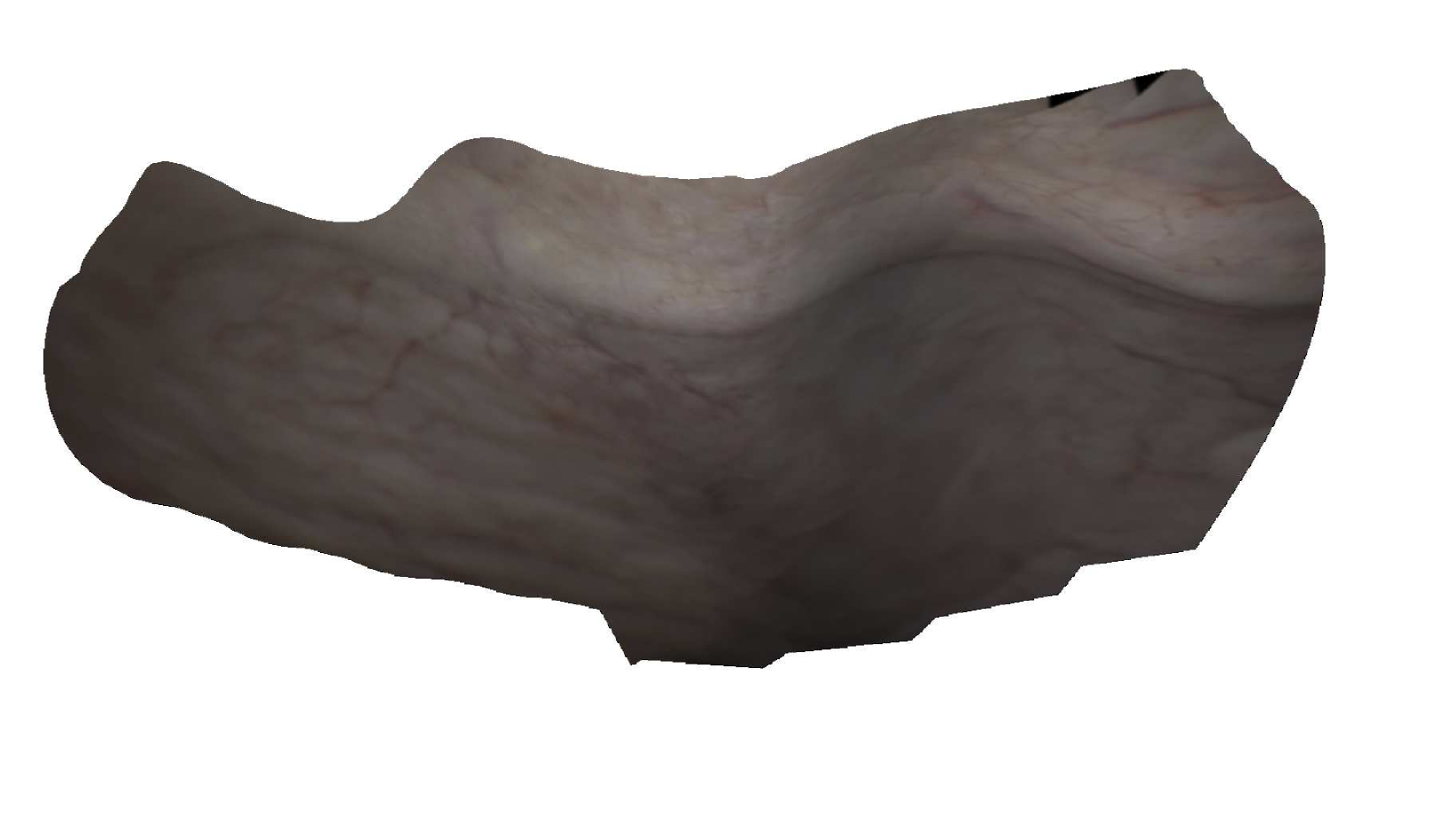} &
\includegraphics[width=0.36\textwidth]{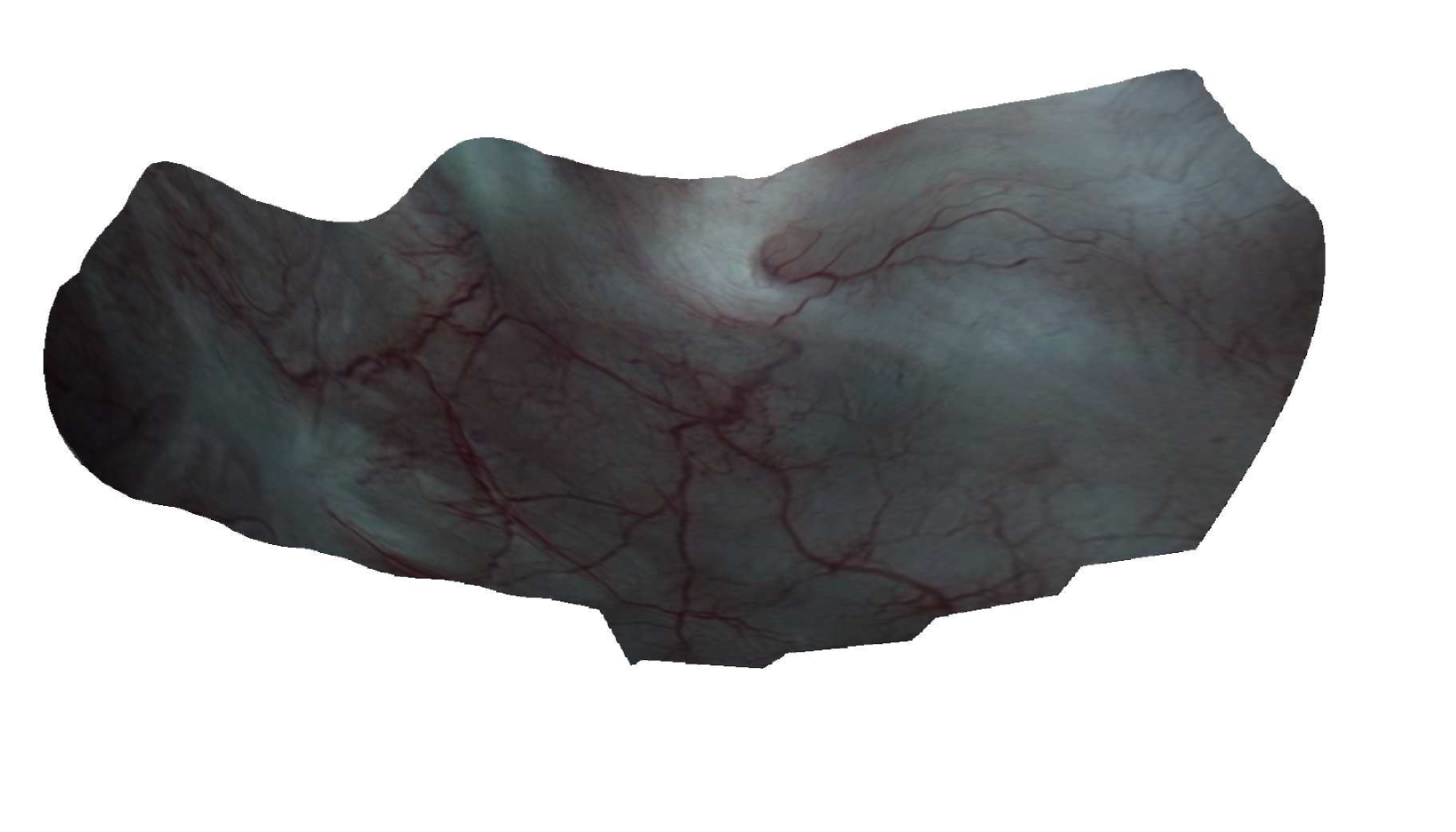} &
\includegraphics[width=0.36\textwidth]{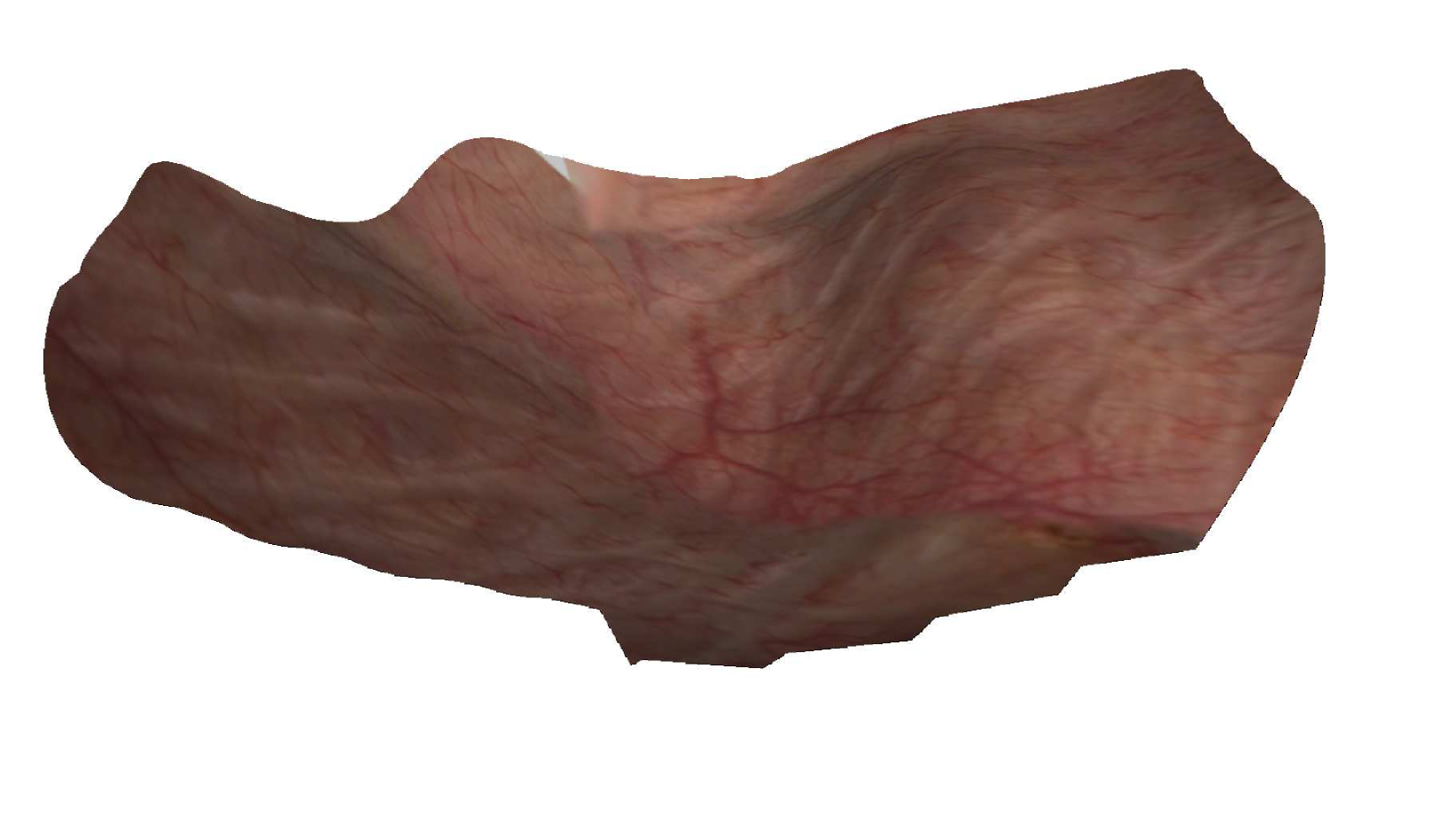} \\[-.4cm]
(d)&(e)&(f)
\end{tabular}
}
\caption{Results obtained for the three numerical phantoms related to
  Fig.~\ref{fig:9}(a,b,c). Top: constructed 3D point cloud (in black)
  and cystoscope trajectory (magenta curve). The point cloud is
  displayed along with the ground truth surfaces (grey mesh) to
  visually assess their overlapping quality. The green dashed curve
  refers to the ground truth cystoscope trajectory. For readability
  reasons, the meshed surface generated from the reconstructed 3D
  points is not shown. Bottom: 3D reconstructed mosaics seen from
  above (for improved readability).  }
\label{fig:10}
\end{figure*}
\subsection{Surface construction results for the simulated data \label{sec:ResNumericalPhantoms}}
The proposed 3D mosaicing method was applied to the three simulated
data sequences. Fig.~\ref{fig:10}(a,b,c) displays the cloud of
reconstructed points $P_{3D}^{i,k}$ for each sequence in the common
coordinate system $\{c_1\}$ together with the trajectory of the
optical center of the cystoscope prototype (the computed trajectory
appears in magenta). The textured surfaces generated from the cloud of
3D points can be found in Fig.~\ref{fig:10}(d,e,f).

The surface construction accuracy is assessed by computing the average
Euclidean distance between the reconstructed 3D points in $\{c_1\}$
and their ground truth locations on the phantom surface. These
distances are ideally equal to 0. The numerical errors obtained for
the three phantoms are 0.79$\pm$0.34 mm, 0.78$\pm$0.33 mm, and
0.80$\pm$0.34 mm, respectively. On Fig. \ref{fig:10}, it is noticeable
that the shape of the reconstructed surface part (containing the black
points) is visually consistent with that of the true surface (grey
mesh). The mean error is small ($<1$ mm), confirming that the
reconstructed extended FOV surfaces remain accurate for both convex
and nonconvex surface parts.
%
%
%
%
\begin{table*}[t!]
\begin{center}
\begin{tabular}{lccccc}
\hline 
{\bf Phantom} & Planar phantom   & Wave phantom & First numerical  & Second numerical  & Third numerical   \\ \vspace*{1mm}
{\bf type} & (Fig.~\ref{fig:5})   & (Fig.~\ref{fig:6}) & phantom (Fig.~\ref{fig:10}(a))  & phantom (Fig.~\ref{fig:10}(b))  & phantom (Fig.~\ref{fig:10}(c))  \\  \cline{2-6}
{\bf Mean} $\epsilon_{3D}^{k-1,k}$ & 0.016$\pm$0.002  & 0.05$\pm$0.01  & 0.021$\pm$ 0.0023 & 0.023$\pm$ 0.0025& 0.023$\pm$0.0027 \\ \cline{2-6}
{\bf Mean} $\epsilon_{2D}^{k-1,k}$ & $0.30\pm$0.29  & 0.85$\pm$0.8  & 0.40$\pm$0.39 & 0.42$\pm$0.37 & 0.43$\pm$0.39 \\ \hline
\end{tabular}
\caption{Registration errors for five phantoms. The mean (over $k$)
  and standard deviation ($mean \pm std$) of criteria
  $\epsilon_{3D}^{k-1,k}$ and $\epsilon_{2D}^{k-1,k}$ are in
  millimetres and pixels, respectively. \label{Tab:RegistrationError}}
\end{center}
\end{table*}
\subsection{Discussion on the simulated data results}
Several conclusions can be drawn. First, the bladder surface of
Fig.~\ref{fig:8} was accurately constructed for video-sequences
containing various human textures. Second, the average 3D error
remains small (less than 1 mm) in all cases. The proposed algorithm
was also able to deal with the pig bladder textures (shown in
Section~\ref{sec:ProtoResultsDiscussion}), which is another indication
of its robustness against bladder texture variations. The use of
mutual information as a similarity measure is the main reason behind
this robustness.

For long video sequences, it may occur that the 3D points
$P_{3D}^{i,k}$ substantially diverge from the ground truth surface.
This is due to the accumulation of 3D registration errors over the
sequence. Error accumulation is a common issue when 2D or 3D
transformations between consecutive frames are sequentially
estimated. This problem was already encountered for 2D mosaicing
algorithms, and efficient solutions were proposed, based on a global
texture discontinuity correction over the processed image
sequence~\cite{Weibel:2012a}.

\section{Assessment of registration accuracy}
\label{sec:3DRegistrationError}
The quantitative results of Sections \ref{sec:ResultsRegisMosaics} and
\ref{sec:ResNumericalPhantoms} are respectively related to the
accuracy of the endoscope displacement and surface estimation. This
section focusses on the registration accuracy between viewpoints.

\subsection{Registration accuracy criteria}
In~\eqref{eq:3DRegistrationCriterion}, $\epsilon_{3D}^{k-1,k}$ defines
the 3D registration error compu\-ted with the known ground truth
transformation $\mathbf{T}_{3D,gt}^{k-1,k}$ and the transformation
$\mathbf{T}_{3D}^{k-1,k}$ estimated with the proposed registration
algorithm. Criterion $\epsilon_{3D}^{k-1,k}$ computes the mean
Euclidean distance in millimetres between the exact and estimated
locations of points $P_{3D}^{k,i}$, displaced in the coordinate system
$\{c_{k-1}\}$ of viewpoint $k-1$:
\begin{align}
\epsilon_{3D}^{k-1,k} = \frac{1}{N}\sum_{i=1}^{N}\left \|
\hat{P}_{3D,gt}^{i,k-1}-\hat{P}_{3D}^{i,k-1}
\right
\| \label{eq:3DRegistrationCriterion}
\end{align}
where 
the displaced locations are respectively denoted by
$\hat{P}_{3D,gt}^{i,k-1}$ and $\hat{P}_{3D}^{i,k-1}$. They are
computed from $\hat{P}_{3D}^{i,k}$ using the ground truth and
estimated transformations $\mathbf{T}_{3D,gt}^{k-1,k}$ and
$\mathbf{T}_{3D}^{k-1,k}$; see~\eqref{eq:matriceTrigidea}. Ideally,
$\epsilon_{3D}^{k-1,k}$ equals 0.

Similarly, in~\eqref{eq:2DRegistrationCriterion}, criterion
$\epsilon_{2D}^{k-1,k}$ computes the mean Euclidean distance in pixels
between the projections in image $I_{k-1}$ of the displaced 3D points
$\hat{P}_{3D,gt}^{i,k-1}$ and $\hat{P}_{3D}^{i,k-1}$:
\begin{align} 
\epsilon_{2D}^{k-1,k}
\hspace*{-.11cm}=\hspace*{-.11cm}\frac{1}{N}\sum_{i=1}^{N}\left \|\frac{1}{\hat{z}_{3D,gt}^{i,k-1}}
\mathbf{K}\hat{P}_{3D,gt}^{i,k-1}-\frac{1}{\hat{z}_{3D}^{i,k-1}}
\mathbf{K}\hat{P}_{3D}^{i,k-1}\right \| 
\label{eq:2DRegistrationCriterion}
\end{align}
$\mathbf{K}\hat{P}_{3D,gt}^{i,k-1}/\hat{z}_{3D,gt}^{i,k-1}$
and $\mathbf{K}\hat{P}_{3D}^{i,k-1}/\hat{z}_{3D}^{i,k-1}$ gather the
coordinates in pixel unit of the ground truth and estimated
point locations, where $\mathbf{K}$ is the projective matrix coding for
the intrinsic camera parameters;
see~\eqref{eq:matricePerspective}. Ideally, $\epsilon_{2D}^{k-1,k}$
also equals 0.

\subsection{Registration results and accuracy discussion}
Table \ref{Tab:RegistrationError} gives the registration errors, \emph{i.e.}, the mean and standard deviation of
the $\epsilon_{3D}^{k-1,k}$  and $\epsilon_{2D}^{k-1,k}$ values computed with the data of viewpoint pairs ($k-1$, $k$) for five phantoms.

The results obtained for the planar phantom correspond to the
inherent accuracy of the algorithm (best possible registration results
obtained with the planarity assumption perfectly fulfilled).  The mean
(0.016 mm) and standard deviation (0.003 mm) of the
$\epsilon_{3D}^{k-1,k}$ values show that in such ideal situation, the
3D registration errors between viewpoints are indeed very small.
The 2D sub-pixel registration error is $\epsilon_{2D}^{k-1,k}=0.30\pm0.29$ pixels for the planar phantom.

The planarity assumption is not fulfilled for the wave phantom: the
field of view of the images is wide and visualizes large wave parts
whose surfaces strongly deviate from a planar shape. In such an
extreme situation, the mean registration error remains small (0.05
mm). Due to these small 3D registration errors, the surface in
Fig.~\ref{fig:6} has a correct global wave shape. Moreover, when
scanning the wave surface, the angle between the 3D cystoscope
prototype and the normal to the local surface takes values ranging in
the whole interval [0-45] degrees. The small standard deviation of
criterion $\epsilon_{3D}^{k-1,k}$ ($0.01$ mm) indicates that the
registration algorithm is robust towards changes
of viewing angle. The mean of the 2D registration error (0.85 pixels)
remains also small.

The 2D and 3D registration errors are quite close for all three
numerical phantoms. The unique difference in the three numerical
phantoms was the human bladder texture used for the data acquisition
simulations (the surface shape and cystoscope trajectory was the same
for all simulations). These results show that the registration
algorithm is robust towards human bladder texture variability (the
mutual information based similarity measure systematically led to
comparable registration accuracy).

The $\epsilon_{3D}^{k-1,k}$ errors measured for the simulated phantoms
are always larger than the $\epsilon_{3D}^{k-1,k}$ score obtained for
the planar phantom (most favorable situation) and smaller than the
$\epsilon_{3D}^{k-1,k}$ score for the wave phantom (most difficult
situation). This is consistent with the fact that the phantom of
Fig.~\ref{fig:5} is exactly planar, the wave phantom is highly
non-planar, whereas the degree of planarity of the numerical phantoms
is intermediate. We would like to stress that the
$\epsilon_{3D}^{k-1,k}$ and $\epsilon_{2D}^{k-1,k}$ scores obtained
for the simulated phantoms are significantly closer to the error
obtained for the plane rather than those of the wave phantom. This is
an indication of the ability of the algorithm to reconstruct bladder
shapes.
\section{Conclusion\label{sec:Conclusion}}
The proposed 3D bladder mosaicing algorithm is guided by 2D image
registration. One advantage of the proposed approach is that no strong
assumption is made on surface shapes. On the one hand, the choice of
some initial surface shape is not required. On the other hand, smooth
surfaces with both convex and nonconvex parts can be accurately
constructed, the latter corresponding to warped bladder
parts. Furthermore, the results obtained for pig bladder data and
human bladder images highlight the robustness of the approach, since
the mutual information based registration can cope with high texture
variability.

An important feature of our endoscope prototype is that the
triangulation geometry for 3D data acquisition and the baseline length
of 3 mm are very close to those of the 3D cystoscopes which might be
built for clinical use. A current trend in endoscopy is the ``chip on
the tip'' technology (\emph{i.e.}, sensors fixed on the distal tip of
endoscopes). In this technology, the laser based solution can be used
for both rigid and flexible cystoscopes (both the CCD matrix and the
diffractive optics can be fixed on the distal tip): the calibration
procedure described in \cite{Ben-Hamadou:2013} and the mosaicing
algorithm proposed here remain unchanged. For rigid cystoscopes, the
calibration and 3D point reconstruction methods are the same for
different lens orientations ($0^\circ$ straight looking lens,
$30^\circ$ forward-oblique looking lens or $70^\circ$ lateral looking
lens).

Except for the optics to be implemented on the cystoscope for 3D point
reconstruction, the modifications to be done for clinical translation
are rather light. This is a key advantage over alternative solutions
which rely on external
devices~\cite{Shevchenko:2012,Agenant:2013}. This should lead to a
drastic reduction of the overall cost of cystoscopy equipment together
with a simplification of the technical installations in the
examination rooms. Actually, the ``chip on the tip'' technology should
allow us to simplify the prototype in future years: the cylindrical
tube of Fig.~\ref{fig:3} cannot be used in clinical situations, and it
will not be used anymore. 

Prior to the construction of a 3D cystoscope usable in clinical
situation, it was required to validate both the measurement principle
and the mosaicing algorithm on phantoms.  Our work is the first which
shows on phantoms with realistic bladder textures that 3D cystoscopy
and mosaicing is feasible while minimally modifying standard
endoscopes. This result is an important step in the process towards
clinical 3D cystoscopy.

The proposed mosaicing algorithm may be improved in several situations.

\indent 1) When cystoscope trajectories are crossing, texture and 3D
surface misalignments will be visible in the mosaic due to small error
accumulation during the mosaicing process. Such crossings have to be
automatically detected so that the surface, texture and color
discontinuities can be corrected. Global 2D mosaic
  correction methods (such as the one proposed in~\cite{Weibel:2012a})
  may be adapted to improve the visual quality of 3D textured
  surfaces, but such approaches will not improve the 3D mosaics
  from the dimensional (accuracy) point of view.

  2) It was shown in recent studies \cite{Ali16b,Ali16a} that optical
  flow can provide an accurate dense correspondence between homologous
  points of bladder images with strong illuminations variations, large
  displacements or perspective changes, blur or weak textures. Thus, a
  perspective of this work is to elaborate an optical flow method with
  energies allowing for simultaneous
  optimization of the vector field and the parameters of the transformation $T_{3D}^{k-1,k}$. 
%
%

  3) The implementation of our algorithm may be improved as well in
  specific cases. When texture features are available and well spread
  over the images, they can be fastly detected \cite{Ali:2013}. Then,
  the distance between homologous texture points can be used as a
  similarity measure for a given $\mathbf{T}_{2D}^{k-1,k}$, instead of
  the mutual information in (14).

When textures are lacking as in Fig.~\ref{fig:1}, the mutual
information has to be used. Second, the processing time can obviously
be reduced by replacing the sequential implementation of image
registration (viewpoints $k-1$ and $k$ for $k=1,2,\ldots$) by a
parallel implementation \emph{e.g.}, on a GPU graphical card. By
coupling fast implementations with future clinical endoscopes based on
the current proof-of-concept, we expect that the image processing time
will remain moderate as compared to the times for acquiring images,
for file transfers, and for archiving and retrieving data.

The registration algorithm was designed for bladder lesion
diagnosis. It can be adapted to minimal invasive surgery in the
abdominal cavity, \emph{e.g.}, for computing extended FOV surfaces of
the stomach. Indeed, the external stomach wall is smooth and contains
textures close to bladder textures. Moreover, a laparoscope associated
to a structured-light projector can be calibrated similar to a
cystoscope. Increasing the FOV and providing 3D information
facilitates a surgical intervention. For this application, real-time
mosaicing must be reached.

\begin{acknowledgements}
  This work was sponsored by the R\'egion Lorraine and the Centre
  National de la Recherche Scientifique (CNRS) under contract PEPS
  Biotechno et Imagerie de la Sant\'e. The authors would also like to
  thank the Centre Hospitalier Universitaire Nancy-Brabois for
  providing pig bladders and Prof. Fran\c{c}ois Guillemin from the
  Institut de Canc\'erologie de Lorraine for his expertise in urology.
\end{acknowledgements}

\bibliographystyle{spmpsci}      

\bibliography{bibliography}

\begin{thebibliography}{10}
\providecommand{\url}[1]{{#1}}
\providecommand{\urlprefix}{URL }
\expandafter\ifx\csname urlstyle\endcsname\relax
  \providecommand{\doi}[1]{DOI~\discretionary{}{}{}#1}\else
  \providecommand{\doi}{DOI~\discretionary{}{}{}\begingroup
  \urlstyle{rm}\Url}\fi

\bibitem{Agenant:2013}
Agenant, T.M., Noordmans, H.J., Koomen, W., Bosch, J.L.: Real-time bladder
  lesion registration and navigation: A phantom study.
\newblock PLoS ONE \textbf{8}(1), e54,348 (2013)

\bibitem{Alcaraz:2007}
Alcaraz, A.: Bladder cancer: Highlights from 2006.
\newblock European Urology Supplements \textbf{6}, 737--744 (2007)

\bibitem{Ali16b}
Ali, S., Daul, C., Galbrun, E., Blondel, W.: Illumination invariant optical
  flow using neighborhood descriptors.
\newblock Computer Vision and Image Understanding \textbf{145}, 95--110 (2016)

\bibitem{Ali16a}
Ali, S., Daul, C., Galbrun, E., Guillemin, F., Blondel, W.: Anisotropic motion
  estimation on edge preserving {R}iesz wavelets for robust video mosaicing.
\newblock {P}attern {R}ecognition \textbf{51}(C), 425--442 (2016)

\bibitem{Ali:2013}
Ali, S., Daul, C., Weibel, T., Blondel, W.: Fast mosaicing of cystoscopic
  images from dense correspondence: combined {SURF and {TV-L1} optical flow
  method}.
\newblock In: 20th IEEE International Conference on Image Processing, pp.
  1291--1295. Melbourne, Australia (2013)

\bibitem{Behrens:2008}
Behrens, A.: Creating panoramic images for bladder fluorescence endoscopy.
\newblock Acta Polytechnica \textbf{48}(3), 50--54 (2008)

\bibitem{Behrens:2009a}
Behrens, A., Stehle, T., Gross, S., Aach, T.: Local and global panoramic
  imaging for fluorescence bladder endoscopy.
\newblock In: 31st Int. Conf. of the IEEE Engineering in Medicine and Biology
  Society, pp. 6690--6693. Minneapolis, USA (2009)

\bibitem{Behrens:2010}
Behrens, A., Uski, M., Stehle, T., Gross, S., Aach, T.: Intensity based
  multi-scale blending for panoramic images in fluorescence endoscopy.
\newblock In: IEEE International Symposium on Biomedical Imaging, pp.
  1305--1308. Rotterdam, Netherlands (2010)

\bibitem{Ben-Hamadou:2010}
Ben-Hamadou, A., Daul, C., Soussen, C., Rekik, A., Blondel, W.: A novel {3D}
  surface construction approach: Application to three-dimensional endoscopic
  data.
\newblock In: 17th IEEE Int. Conf. on Image Processing, pp. 4425--4428. Hong
  Kong (2010)

\bibitem{Ben-Hamadou:2013}
Ben-Hamadou, A., Soussen, C., Daul, C., blondel, W., Wolf, D.: Flexible
  calibration of structured-light systems projecting point patterns.
\newblock Computer Vision and Image Understanding \textbf{117}(10), 1468--1481
  (2013)

\bibitem{Bergen16}
Bergen, T., Wittenberg, T.: Stitching and surface reconstruction from
  endoscopic image sequences: A review of applications and methods.
\newblock IEEE Journal of Biomedical and Health Informatics \textbf{20}(1),
  304--321 (2016)

\bibitem{Bergen13}
Bergen, T., Wittenberg, T., M\"unzenmayer, C., Chen, C.C.G., Hager, G.D.: A
  graph-based approach for local and global panorama imaging in cystoscopy.
\newblock In: Proc. {SPIE}, {M}edical {I}maging: Image-Guided Procedures,
  Robotic Interventions, and Modeling, vol. 8671. Lake Buena Vista, Florida,
  USA (2013)

\bibitem{blender:2014}
Blender-Online-Community: {B}lender - a 3{D} modelling and rendering package
  (2014).
\newblock \urlprefix\url{{www.blender.org}}

\bibitem{Carroll:2009}
Carroll, R., Seitz, S.: Rectified surface mosaics.
\newblock International Journal on Computer Vision \textbf{85}(3), 307--315
  (2009)

\bibitem{Chan2003}
Chan, M., Lin, W., Zhou, C., Qu, J.Y.: Miniaturized three-dimensional
  endoscopic imaging system based on active stereovision.
\newblock Applied Optics \textbf{42}(10), 1888--1898 (2003)

\bibitem{daul:2000}
Daul, C., R{\"o}sch, R., Claus, B.: Building a color classification system for
  textured and hue homogeneous surfaces: system calibration and algorithm.
\newblock Machine Vision and Applications \textbf{12}(3), 137--148 (2000)

\bibitem{Hernandez:2010}
Hern\'andez-Mier, Y., Blondel, W.C.P.M., Daul, C., Wolf, D., Guillemin, F.:
  Fast construction of panoramic images for cystoscopic exploration.
\newblock Computerized Medical Imaging and Graphics \textbf{34}(7), 579--592
  (2010)

\bibitem{Holzbeierlein:2004}
Holzbeierlein, J., Lopez-Corona, E., Bochner, B., Herr, H., Donat, S., Russo,
  P., Dalbagni, G., Sogani, P.: Partial cystectomy: a contemporary review of
  the memorial {S}loan-kettering cancer center experience and recommendations
  for patient selection.
\newblock Journal of Urology \textbf{172}(3), 878--881 (2004)

\bibitem{Kaufman:2008}
Kaufman, A., Wang, J.: {3D} surface reconstruction from endoscopic videos.
\newblock In: L.~Linsen, H.~Hagen, B.~Hamann (eds.) Visualization in Medicine
  and Life Sciences, Mathematics and Visualization, pp. 61--74. Springer Berlin
  Heidelberg (2008)

\bibitem{Kazhdan:2006}
Kazhdan, M., Bolitho, M., Hoppe, H.: Poisson surface reconstruction.
\newblock In: Fourth Eurographics symposium on Geometry processing, pp. 61--70.
  Cagliari, Italy (2006)

\bibitem{Miranda:2004}
Miranda, R., Hernandez-Mier, Y., Daul, C., Blondel, W., Wolf, D.: Mosaicing of
  medical video-endoscopic images : data quality improvement and algorithm
  testing.
\newblock In: International Conference on Electrical and Electronics
  Engineering (ICEEE) and Tenth Conference on Electrical Engineering (CIE).
  Acapulco Guerrero, Mexico (2004)

\bibitem{Miranda2008}
Miranda-Luna, R., Daul, C., Blondel, W., Hern\'andez-Mier, Y., Wolf, D.,
  Guillemin, F.: Mosaicing of bladder endoscopic image sequences: Distortion
  calibration and registration algorithm.
\newblock {IEEE} Transactions on {B}iomedical {E}ngineering \textbf{55}(2),
  541--553 (2008)

\bibitem{Mountney2009}
Mountney, P., Yang, G.Z.: Dynamic view expansion for minimally invasive surgery
  using simultaneous localization and mapping.
\newblock In: Int. Conf. of the IEEE Engineering in Medicine and Biology
  Society, pp. 1184--1187. Minneapolis, USA (2009)

\bibitem{Nelder:1965}
Nelder, J.A., Mead, R.: A simplex method for function minimization.
\newblock The computer journal \textbf{7}(4), 308--313 (1965)

\bibitem{Penne2009}
Penne, J., H{\"o}ller, K., St{\"u}rmer, M., Schrauder, T., Schneider, A.,
  Engelbrecht, R., Feu{\ss}ner, H., Schmauss, B., Hornegger, J.: Time-of-flight
  3-{D} endoscopy.
\newblock Medical Image Computing and Computer-Assisted Intervention pp.
  467--474 (2009)

\bibitem{Pluim:2003}
Pluim, J., Maintz, J., Viergever, M.: Mutual-information-based registration of
  medical images: a survey.
\newblock IEEE Trans. on Medical Imaging \textbf{22}(8), 986--1004 (2003)

\bibitem{Shevchenko:2012}
Shevchenko, N., Fallert, J., Stepp, H., Sahli, H., Karl, A., Lueth, T.: A high
  resolution bladder wall map: Feasibility study.
\newblock In: 34th Int. Conf. of the IEEE Engineering in Medicine and Biology
  Society, pp. 5761--5764. San Diego, CA (2012)

\bibitem{Soper2012}
Soper, T., Porter, M., Seibel, E.J.: Surface mosaics of the bladder
  reconstructed from endoscopic video for automated surveillance.
\newblock {IEEE} Transactions on {B}iomedical {E}ngineering \textbf{59}(6),
  1670--1680 (2012)

\bibitem{Weibel2012b}
Weibel, T., Daul, C., Wolf, D., R\"osch, R.: Contrast-enhancing seam detection
  and blending using graph cuts.
\newblock In: 21st International Conference on Pattern Recognition, pp.
  2732--2735. Tsukuba, Japan (2012)

\bibitem{Weibel:2012a}
Weibel, T., Daul, C., Wolf, D., R\"osch, R., Guillemin, F.: Graph based
  construction of textured large field of view mosaics for bladder cancer
  diagnosis.
\newblock Pattern Recognition \textbf{45}(12), 4138--4150 (2012)

\bibitem{Wu:2010}
Wu, C., Narasimhan, S., Jaramaz, B.: A multi-image shape-from-shading framework
  for near-lighting perspective endoscopes.
\newblock International Journal of Computer Vision \textbf{86}(2-3), 211--228
  (2010)

\bibitem{Wu:2007}
Wu, C.H., Sun, Y.N., Chang, C.C.: Three-dimensional modeling from endoscopic
  video using geometric constraints via feature positioning.
\newblock IEEE Transactions on Biomedical Engineering \textbf{54}(7),
  1199--1211 (2007)

\end{thebibliography}
\end{document}